\documentclass[11pt]{article}

\usepackage[preprint]{acl}

\usepackage{times}
\usepackage{latexsym}

\usepackage[T1]{fontenc}

\usepackage[utf8]{inputenc}

\usepackage{microtype}

\usepackage{inconsolata}

\usepackage{graphicx}

\usepackage{algorithm}
\usepackage{algpseudocode}
\usepackage{amsmath}
\usepackage{amsfonts}
\usepackage{amsthm}
\usepackage{amssymb}
\usepackage{booktabs}
\usepackage{multirow}
\usepackage{multicol}
\usepackage{makecell}
\usepackage{enumerate}
\usepackage{enumitem}
\usepackage{caption}
\usepackage{xcolor}
\usepackage{float}
\usepackage{graphicx}
\usepackage{color}
\usepackage{xspace}
\usepackage{tikz}
\usetikzlibrary{arrows.meta,positioning,decorations.pathreplacing}
\usepackage{subcaption}
\usepackage{placeins}
\usepackage[table]{xcolor}
\usepackage[most]{tcolorbox}
\tcbset{
  colback=gray!5,
  colframe=gray!70,
  boxrule=0.5pt,
  arc=6.5pt,
  left=6pt,right=6pt,top=4pt,bottom=4pt,
}
\newtcolorbox{highlightbox}{}

\usepackage{booktabs}
\usepackage{tabularx}
\usepackage{array}

\newcommand{\model}{\textsc{CoMap}\xspace}

\usepackage{todonotes}

\newcommand{\qwenicon}{\raisebox{-0.3\height}{\includegraphics[height=3.5ex]{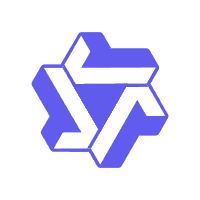}}}

\newcommand{\deepseekicon}{%
  \raisebox{-0.18\height}{\includegraphics[height=2.13ex]{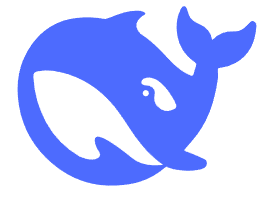}}%
}

\newcommand{\chatgpticon}{%
  \raisebox{-0.20\height}{\includegraphics[height=2.20ex]{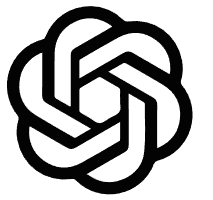}}%
}

\DeclareRobustCommand{\prompticon}{%
  \raisebox{-0.18em}{\includegraphics[height=1.05em]{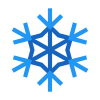}}%
}

\DeclareRobustCommand{\trainicon}{%
  \raisebox{-0.18em}{\includegraphics[height=1.05em]{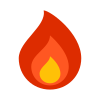}}%
}
\usepackage{xcolor}
\usepackage{pifont}

\usepackage{hyperref}
\usepackage{fontawesome5}
\hypersetup{
    colorlinks=true,
    urlcolor=blue
}

\title{\textsc{CoMap}: Co-Evolving World Models and Agent Policies for LLM Agents}

\author{
     Youwei Liu$^{1}$, ~~Jian Wang$^{2 \dagger}$, ~~Hanlin Wang$^{3}$, ~~Wenjie Li$^{3}$ \\
     $^1$ Central South University ~~ $^2$ College of Computer Science, Sichuan University \\
     $^3$ Department of Computing, The Hong Kong Polytechnic University \\
     \texttt{loyiv5477@gmail.com  ~ wangjian51@scu.edu.cn}
     \\ \texttt{hanlin-henry.wang@connect.polyu.hk ~ cswjli@comp.polyu.edu.hk}
}

\begin{document}
\maketitle

\renewcommand{\thefootnote}{$\dagger$}
\footnotetext[1]{~Corresponding author.}
\setcounter{footnote}{0}
\renewcommand{\thefootnote}{\arabic{footnote}}

\begin{abstract}
World models enable language agents to anticipate environment dynamics and evaluate candidate actions before execution, yet they are typically frozen after training, leaving them unable to adapt to the shifting on-policy distributions induced by evolving agents. 
Meanwhile, agent-improvement methods often depend on external rewards or verifiers that may be unavailable in realistic interactive environments.
We introduce \textsc{CoMap}, a framework that co-evolves textual world models and agent policies through closed-loop interaction. 
At each decision step, the world model predicts future environment feedback for candidate actions, while the agent performs future-aware reflection by assessing the reliability of these predictions and refining its action accordingly. 
The resulting on-policy trajectories are then used to improve the world model through self-distillation, enabling it to continually adapt to the agent's evolving interaction distribution. 
Across embodied task planning, web navigation, and tool-use benchmarks, \textsc{CoMap} consistently outperforms competitive baselines (e.g., a 16.75\% relative improvement with Qwen3-4B). 
Further analyses show that this co-evolutionary process improves both world-model capability and long-horizon decision-making.
Our code and data are available at: \noindent\faGithub\ \url{https://github.com/loyiv/CoMAP}.

\end{abstract}

\section{Introduction}

 \begin{figure}[th!]
    \centering
    \includegraphics[width=\linewidth]{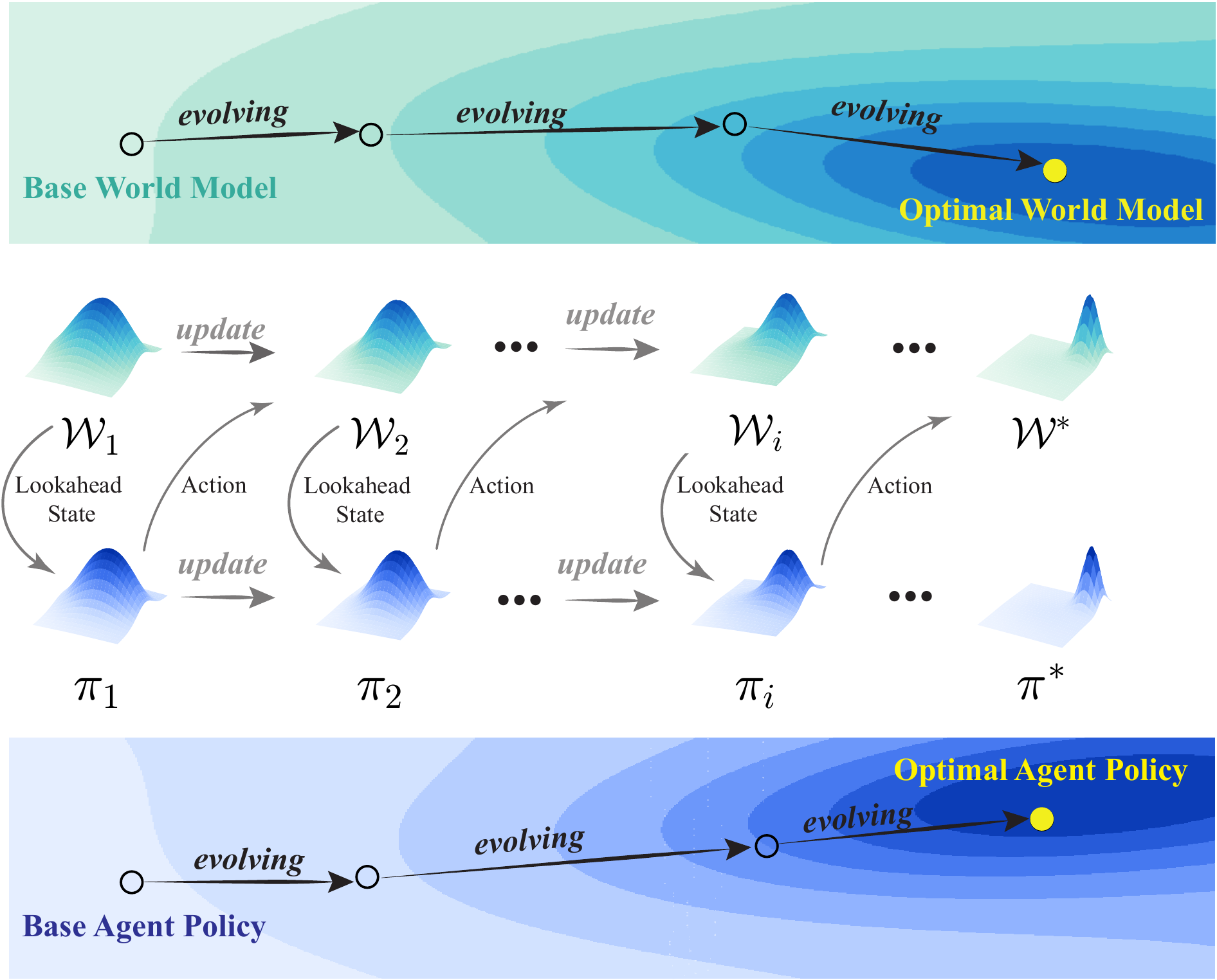}
    \caption{Conceptual illustration of the co-evolution of world models and agent policies for LLM Agents.}
    \label{fig:coevolving_overview}
    \vspace{-6pt}
\end{figure}

The emergence of large language models (LLMs) has enabled intelligent agents to reason, plan, and interact across a wide range of digital and physical environments~\citep{yao2023react,wang2024survey,li2024embodied}.
Recent studies suggest that equipping LLM-based agents with world-modeling capabilities, namely the ability to anticipate environment state dynamics with respect to candidate actions, can substantially improve their decision-making by enabling agents to evaluate future outcomes before execution~\citep{hafner2023mastering,sun2024enhancing,chae2025web,liu2026itp}. 
However, existing textual world models are typically fixed after training or construction~\cite{zhang2025agent,li2025word}. As a result, they struggle to adapt to the on-policy state-action distributions induced by the agent's own evolving behavior, especially in long-horizon interactive tasks.
In parallel, agent-evolving methods improve policies through continual optimization, but often depend on external rewards, human-written rubrics, or task-specific verifiers~\citep{sheng2026rlcer}. 
This motivates a central question: \textit{Can the internal future predictions of a world model serve as a self-improving signal that simultaneously refines the agent policy and adapts the world model itself, without requiring external annotations?}

We argue that world modeling and policy learning should be treated as a coupled co-evolutionary process. 
The key challenge is that the two components interact under a non-stationary distribution. A fixed world model may provide inaccurate or poorly calibrated predictions when the agent reaches states that differ from those observed during training. Conversely, a weak policy may generate uninformative trajectories, limiting the world model's exposure to diverse and task-critical transitions. 
Improving either component in isolation can therefore create a bottleneck: unreliable future predictions constrain policy improvement, while suboptimal policies restrict the data available for world-model adaptation. 
As illustrated in Figure~\ref{fig:coevolving_overview}, an effective agent should instead form a closed loop in which the world model provides forward-looking guidance for action decision-making, and the agent's on-policy interactions provide fresh evidence for updating the world model.

With this in mind, we propose \textbf{Co}-evolving world \textbf{M}odels and \textbf{A}gent \textbf{P}olicies (\textbf{\model}), a closed-loop framework that enables the world model and the agent policy to improve each other through interaction. \model consists of two mutually reinforcing loops. 
In the decision loop, the agent first drafts a candidate action from the current state, and the world model yields future feedback, such as the task-relevant consequences of that action or the next possible observation. 
The agent then estimates its reliability and performs \emph{future-aware reflection} to refine this action. 
Similarly, the resulting on-policy trajectories are used to construct \emph{self-distillation} targets for the world model, aligning its predictions with the dynamic state-action distribution induced by the current policy. 
As the policy improves, it reaches more complex, informative states; as the world model adapts, it provides more faithful predictions, thereby further guiding the agent's long-horizon decisions.

We evaluate \model across widely used benchmarks~\citep{yao2022webshop, Wang2022ScienceWorld,ALFWorld20,guo-etal-2024-stabletoolbench}.
Extensive experiments demonstrate that \model achieves significantly higher task success rates over competitive baselines. Further analyses demonstrate that the co-evolutionary process improves the world model's ability to predict environment state dynamics, which in turn provides more reliable foresight for action decision-making.

In summary, our contributions are as follows:
\begin{itemize}[leftmargin=*]
    \item We identify a key limitation of existing world-model-based language agents: textual world models are often static and therefore fail to adapt to the on-policy interaction distributions produced by evolving agent policies.
    
    \item We propose \model, a closed-loop framework that co-evolves the world model and the agent policy by leveraging future-aware reflection to refine decisions, with on-policy self-distillation to improve the world model.
    
    \item Extensive experiments on embodied task planning, web navigation, and tool-use benchmarks show that \model consistently outperforms competitive baselines, providing insightful evidence that our co-evolution enhances world modeling and yields effective policies for complex tasks.
\end{itemize}

\section{Preliminaries}
\label{sec:prelim}

\paragraph{Problem Formulation.}

\begin{figure*}[t!]
    \centering
    \includegraphics[width=\textwidth]{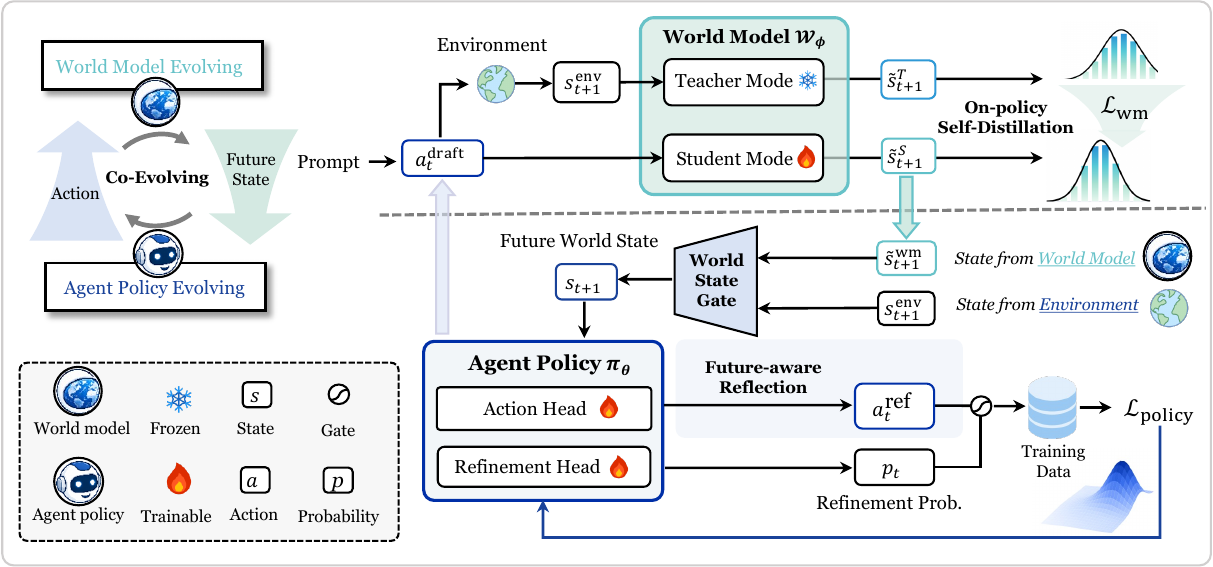}
    \caption{Overview of the proposed \model framework. The textual world model and the agent policy co-evolve through three key mechanisms: on-policy self-distillation of the world model, future-aware policy reflection, and dynamic interaction between the model and the policy.}
    \label{fig:method_overview}
    \vspace{-6pt}
\end{figure*}


The decision-making process of an LLM agent can be formulated as a partially observable Markov decision process (POMDP), denoted by
$\mathcal{M}=(\mathcal{S},\mathcal{A},\mathcal{O},\mathcal{T},\mathcal{R})$.
Here, $\mathcal{S}$ is the environment state space, $\mathcal{A}$ is the action space, and $\mathcal{O}$ is the observation space.
$\mathcal{T}$ represents the state transition function, and $\mathcal{R}$ denotes the reward function that measures task performance.
At each time step $t$, the environment has an underlying state $s_t\in\mathcal{S}$, while the agent receives a partial observation $o_t\in\mathcal{O}$.
Since the true state is not directly accessible, the agent conditions its decision on the interaction history
$h_t=(o_1,a_1,\ldots,a_{t-1},o_t)$, then the agent policy $\pi_\theta$ generates an action $a_t\sim\pi_\theta(\cdot\mid h_t)$.
After executing $a_t$, the environment transits to the next state according to $s_{t+1}\sim\mathcal{T}(\cdot\mid s_t,a_t)$.

\paragraph{Textual World Models.}
We introduce textual world models to provide the agent with additional in-context information in the form of predicted future states~\cite{zhang2025agent,li2025word}.
In text-based environments, the agent interacts through natural-language observations, and the available observation history can be viewed as a textual state representation.
For simplicity, we denote this textual state as $s_t$.
Given $s_t$ and a candidate action $a_t$, an autoregressive LLM can serve as a textual world model $\mathcal{W}_\phi$ by modeling the transition distribution
$p_\phi(s_{t+1}\mid s_t,a_t)$.

With a world model, we follow \citet{liu2026itp} to reformulate the agent's POMDP decision process.
The agent policy first generates a draft action
$a_t^{\mathrm{draft}}\sim\pi_\theta(\cdot\mid h_t)$, then the world model imagines its one-step consequence:
$\hat{s}_{t+1}=\mathcal{W}_\phi(s_t,a_t^{\mathrm{draft}})$.
In this way, the predicted future state is then fed back to the agent policy, yielding a foresight-augmented trajectory $h_t^{+}=(h_t,a_t^{\mathrm{draft}},\hat{s}_{t+1})$ for the final policy decision.
\section{Method}
\label{sec:method}

We propose \textbf{Co}-evolving world \textbf{M}odels and \textbf{A}gent \textbf{P}olicies (\textbf{\model}), a novel framework in which an agent policy and a world model improve each other over time.
Figure~\ref{fig:method_overview} shows the overview of our \model framework.

\subsection{World Model-guided Decision-Making}

\paragraph{Future State Imagination.}
Let $\pi_\theta$ denote the agent policy and $\mathcal{W}_\phi$ denote the world model. 
At time step $t$, given the current state $s_t$, the agent policy first proposes a draft action:
\begin{equation}
    a_t^{\mathrm{draft}} \sim \pi_\theta(\cdot \mid s_t).
\end{equation}
The world model then performs a lookahead imagination and predicts the next state as:
\begin{equation}
    \hat{s}_{t+1}=\mathcal{W}_{\phi}(s_t,a_t^{\mathrm{draft}}).
\end{equation}
This predicted state serves as textual future feedback that describes the likely consequence of the draft action. 
This imagined feedback is then incorporated into the agent policy to refine its decision-making when necessary.

\paragraph{Future-aware Reflection.}
We use the future state to construct self-refinement signals for policy.
Given the future state $\hat{s}_{t+1}$, the policy then conditions on $(s_t,a_t^{\mathrm{draft}},\hat{s}_{t+1})$ to generate a reflected action $a_t^{\mathrm{ref}}$ and a refinement probability $p_t$.
The resulting data $\mathcal{D}_{\mathrm{ref}}
=\{(s_t,a_t^{\mathrm{draft}},\hat{s}_{t+1},a_t^{\mathrm{ref}},p_t)\}$
is used for agent policy-side evolving.
The policy reasons over the current state, the draft action, and the imagined state to produce the next action:
\begin{equation}
a_t \sim \pi_\theta(\cdot \mid s_t, a_t^{\mathrm{draft}}, \hat{s}_{t+1}).
\end{equation}

\subsection{Initialized Training}
Before co-evolving, we initialize the agent policy and the world model with lightweight supervision. This provides a usable starting point while leaving the main improvement to on-policy co-evolution.

\paragraph{Agent Policy and World Model Warm-up.}
Given expert demonstrations $\mathcal{D}_{\mathrm{exp}}=\{(s_t,a_t^{\mathrm{exp}})\}$ and real on-policy transitions from rolling out in the environment $\mathcal{D}_{\mathrm{roll}}=\{(s_t,a_t,s_{t+1}^{\mathrm{env}})\}$, we warm up both the agent policy and the textual world model with supervised fine-tuning as follows:
\begin{align}
\mathcal{L}_{\mathrm{policy}}^{\mathrm{exp}}
&= -\sum\log p_\theta(a_t^{\mathrm{exp}}\mid s_t),
\\
\mathcal{L}_{\mathrm{WM}}
&= -\sum\log p_\phi(s_{t+1}^{\mathrm{env}}\mid s_t,a_t).
\end{align}

\paragraph{Reflection-mode Initialization.}
To initialize the reflection mode, we construct training signals based on the outcomes of future rollouts from expert demonstrations.
For each state $s_t$, the policy first proposes a draft action $a_t^{\mathrm{draft}}$, which is evaluated alongside the corresponding expert action.
We restore the environment to $s_t$ and roll out two parallel trajectory branches, one initiated by the draft action and the other by the expert action, both under the same continuation policy.
If the expert branch achieves a higher success rate or a larger cumulative future return, we assign a positive refinement label $y_t^{\mathrm{ref}}=1$ and designate the expert action as the target.
Otherwise, we set $y_t^{\mathrm{ref}}=0$ and retain the draft action.
Note that draft actions equivalent to the expert action after canonicalization are inherently treated as non-refinement cases.
Finally, taking the future-conditioned observation $x_t^{\mathrm{ref}}=[s_t; a_t^{\mathrm{draft}}; \hat{s}_{t+1}]$ as input, we train an action head and a refinement head (see Figure~\ref{fig:method_overview}) to jointly predict the refined action $a_t^{\mathrm{ref}}$ and its refinement probability $p_t$.

\subsection{Co-Evolving}
Co-evolving aims to jointly improve the agent policy and the textual world model through two coupled learning loops.
We propose \emph{on-policy self-distillation for the world model}, where a ``student'' world model is trained on real executed transitions and regularized by a ``teacher'' model.
On the agent policy side, we leverage \emph{future-aware reflection} to refine its draft action based on a gated future-state signal.
These two loops reinforce each other: the evolving policy expands the on-policy transition distribution for world-model learning, while the evolving world model supplies increasingly reliable lookahead states for policy refinement. 

\paragraph{Evolving World Models via On-policy Self-distillation.}
After the initial warm-up, the world model is instantiated with a student mode
$\mathcal{W}_{\phi}^{S}$ and a teacher mode
$\mathcal{W}_{\bar{\phi}}^{T}$.
For each on-policy transition $(s_t,a_t,s_{t+1}^{\mathrm{env}})$, the realistic next state
$s_{t+1}^{\mathrm{env}}$ is used as privileged information for the teacher mode.
The student first generates an on-policy next-state rollout $\hat{s}_{t+1}^{S}\sim p_{\phi}^{S}(\cdot\mid s_t,a_t)$, while the teacher observes the same input together with $s_{t+1}^{\mathrm{env}}$ and provides a soft target for this student rollout.
This privileged teacher mode is used during training to provide transition-level supervision and reliability estimation.

Let $L_t=|\hat{s}_{t+1}^{S}|$ be the length of the student-generated next-state sequence.
At token position $j$, the student distribution is 
$q_{t,j}^{S}=p_{\phi}^{S}(\cdot\mid s_t,a_t,\hat{s}_{t+1,<j}^{S})$, 
while the teacher distribution is 
$q_{t,j}^{T}=p_{\bar{\phi}}^{T}(\cdot\mid s_t,a_t,s_{t+1}^{\mathrm{env}},\hat{s}_{t+1,<j}^{S})$.
Here, both distributions are evaluated on the same student-generated prefix $\hat{s}_{t+1,<j}^{S}$, but the teacher additionally observes the privileged state $s_{t+1}^{\mathrm{env}}$. 
The world-model self-distillation loss is defined as: 
\begin{equation}
\mathcal{L}_{\mathrm{WMSD}}
=\frac{1}{L_t}\sum_{j=1}^{L_t}
D\!\left(\operatorname{sg}(q_{t,j}^{T})\Vert q_{t,j}^{S}\right),
\end{equation}
which averages the token-level divergence between the teacher and student distributions along the student-generated rollout.
Here, $D(\cdot\Vert\cdot)$ denotes a distribution divergence such as the KL divergence, and $\operatorname{sg}(\cdot)$ stops gradients through the teacher.
The world model is finally trained as follows:
\begin{equation}
\begin{aligned}
\mathcal{L}_{\mathrm{WM}}
=
\mathbb{E}_{\tau_t\sim\mathcal{D}_{\mathrm{roll}}}
\Big[
&-\log p_{\phi}^{S}(s_{t+1}^{\mathrm{{env}}}\mid s_t,a_t) \\
&+\eta\mathcal{L}_{\mathrm{WMSD}}
\Big].
\end{aligned}
\end{equation}
After each student update, the teacher is updated by an exponential moving average, i.e., $\bar{\phi}\leftarrow\mu\bar{\phi}+(1-\mu)\phi$.

\paragraph{Evolving Agent Policies via Future-aware Reflection.}
For agent-policy evolving, the student world model first provides a lookahead future signal for the draft action:
$\hat{s}_{t+1}^{S}=\mathcal{W}_{\phi}^{S}(s_t,a_t^{\mathrm{draft}})$.
The policy then conditions on
$x_t^{\mathrm{ref}}=[s_t;a_t^{\mathrm{draft}};\hat{s}_{t+1}]$
to generate a reflected action $a_t^{\mathrm{ref}}$ and a refinement probability $p_t$.
Given the gated online reflection data $\mathcal{D}_{\mathrm{ref}}$, we set
$a_t^{\mathrm{tgt}}=a_t^{\mathrm{ref}}$ when the refinement is accepted, and otherwise keep
$a_t^{\mathrm{tgt}}=a_t^{\mathrm{draft}}$.
The policy objective is:
\begin{equation}
\begin{aligned}
\mathcal{L}_{\mathrm{policy}}
=
&\mathcal{L}_{\mathrm{policy}}^{\mathrm{exp}}
+
\mathcal{L}_{\mathrm{policy}}^{\mathrm{ref}}
\\
&+
\alpha\,
\mathbb{E}_{\mathcal{D}_{\mathrm{ref}}}
\left[
-\log p_\theta(a_t^{\mathrm{tgt}}\mid x_t^{\mathrm{ref}})
\right]
\\
&+
\beta\,
\mathbb{E}_{\mathcal{D}_{\mathrm{ref}}}
\left[
\operatorname{BCE}(p_t,y_t^{\mathrm{ref}})
\right].
\end{aligned}
\end{equation}
Here, $\mathcal{L}_{\mathrm{policy}}^{\mathrm{exp}}$ anchors the draft mode with expert imitation, while
$\mathcal{L}_{\mathrm{policy}}^{\mathrm{ref}}$ denotes the reflection mode with trajectory rollouts.
BCE is short for the binary cross-entropy.

\paragraph{Co-Evolving Training.}
The two learning loops are coupled through online interaction: the policy drafts an action, the world model predicts its future state, the policy reflects and executes an action, and the resulting transition is used to update the world model.
This closed loop can introduce two kinds of noise: inaccurate early future states may mislead reflection, and unnecessary refinements may harm already good draft actions.
We therefore use a \emph{world-state gate} to decide whether $\hat{s}_{t+1}$ comes from the student prediction $\hat{s}_{t+1}^{S}$ or the real next state $s_{t+1}^{\mathrm{env}}$ during training, and an \emph{action gate} to decide whether the reflected action is adopted or the draft action is kept.
The accepted and rejected cases form $\mathcal{D}_{\mathrm{ref}}$ with labels $y_t^{\mathrm{ref}}$.
With the co-evolving training, the agent policy learns from gated future-aware reflection, while the world model learns from the induced on-policy transitions.

\paragraph{Inference.}
At inference time, the next environment state and the privileged teacher are unavailable before action execution. 
Therefore, the agent uses the student world model to imagine the future state, and produces $(r_t,a_t^{\mathrm{ref}},p_t)$ through future-aware reflection. 
The ultimate action is yielded by the action gate: the agent executes $a_t^{\mathrm{ref}}$ only when the action gate chooses the reflected action; otherwise, it falls back to the draft action $a_t^{\mathrm{draft}}$. 
Details of implementation and mechanism clarification are provided in Appendix~\ref{app:mechanism_clarification}.

\section{Experiments}
\label{sec:experiment}

\subsection{Experimental Settings}

\begin{table*}[th!]
\centering
\resizebox{\textwidth}{!}{
\begin{tabular}{llccccccc}
\toprule
\multirow{2}{*}{\textbf{Type}} 
& \multirow{2}{*}{\textbf{Method}} 
& \multicolumn{2}{c}{\textbf{ALFWorld (AW)}} 
& \multicolumn{2}{c}{\textbf{ScienceWorld (SW)}}
& \multirow{2}{*}{\makecell{\textbf{WebShop} \\ \textbf{(WS)}}} 
& \multirow{2}{*}{\makecell{\textbf{StableToolBench} \\ \textbf{(STB)}}}
& \multirow{2}{*}{\textbf{Avg.}} \\
\cmidrule(lr){3-4} 
\cmidrule(lr){5-6} 
& & \textit{Seen} & \textit{Unseen}  & \textit{Seen} & \textit{Unseen} &  &  & \\
\midrule

\rowcolor{gray!12}
\multicolumn{9}{c}{\textit{\textbf{API-Called Models}}} \\
\midrule

\multirow{2}{*}{\textit{GPT-5.4} ~\protect~\chatgpticon}
& \prompticon\ ReAct           
& 83.20 & 80.60 & 58.10 & 53.40 & 64.90 & 66.80 & 67.83 \\
& \prompticon\ Imagine-and-Act 
& 86.90 & 84.70 & 60.70 & 55.80 & 67.40 & 69.20 & 70.78 \\
\specialrule{0.25pt}{1pt}{1pt}

\multirow{2}{*}{\textit{DeepSeek-V4-Flash}~\protect~\deepseekicon}
& \prompticon\ ReAct           
& 67.80 & 64.50 & 49.60 & 45.30 & 55.10 & 54.80 & 56.18 \\
& \prompticon\ Imagine-and-Act 
& 71.20 & 68.30 & 52.70 & 48.60 & 58.40 & 58.30 & 59.58 \\

\specialrule{0.25pt}{1pt}{1pt}

\multirow{2}{*}{\textit{DeepSeek-V4-Pro}~\protect~\deepseekicon}
& \prompticon\ ReAct           
& 82.40 & 80.20 & 66.50 & 62.80 & 70.20 & 62.50 & 70.77 \\
& \prompticon\ Imagine-and-Act 
& 86.10 & 83.90 & 69.80 & 66.30 & 73.50 & 65.20 & 74.13 \\

\midrule

\rowcolor{gray!12}
\multicolumn{9}{c}{\textit{\textbf{Backbone Model: Qwen3-4B}}~\protect\qwenicon} \\
\midrule

\multirow{4}{*}{Base Agent Planning}
& \prompticon\ ReAct                 
& 14.29 & 12.50 & 6.32  & 5.44  & 13.85 & 21.50 & 12.32 \\

& \hspace{0.6em}$\hookrightarrow$ \trainicon\ w/ SFT               
& 62.86 & 45.80 & 39.50 & 34.20 & 36.80 & 35.50 & 42.44 \\

& \prompticon\ Imagine-and-Act                 
& 20.00 & 18.20 & 8.60  & 7.90  & 16.40 & 24.00 & 15.85 \\

& \hspace{0.6em}$\hookrightarrow$ \trainicon\ w/ SFT               
& 65.71 & 48.40 & 42.30 & 36.70 & 40.10 & 39.00 & 45.37 \\

\specialrule{0.25pt}{1pt}{1pt}

\multirow{3}{*}{Test-time Evolving}
& \prompticon\ Reflexion 
& 21.43 & 19.20 & 8.90  & 8.40  & 17.80 & 23.50 & 16.54 \\

& \prompticon\ RAP       
& 24.29 & 22.10 & 10.80 & 12.50 & 14.90 & 25.00 & 18.27 \\

& \prompticon\ ITP$_\text{I}$   
& 31.43 & 28.70 & 13.60 & 14.20 & 19.80 & 30.00 & 22.95 \\

\specialrule{0.25pt}{1pt}{1pt}

\multirow{3}{*}{World Modeling}
& \trainicon\ WKM       
& 62.14 & 44.90 & 43.70 & 37.50 & 45.80 & 35.00 & 44.84 \\

& \trainicon\ IWM       
& 64.29 & 47.20 & 45.20 & 40.60 & 44.70 & 37.50 & 46.58 \\

& \trainicon\ ITP$_\text{R}$   
& 67.15 & 48.54 & 49.85 & 43.20 & 52.35 & 48.00 & 51.52 \\

\specialrule{0.25pt}{1pt}{1pt}

\rowcolor{blue!8}
Co-evolving
& \trainicon\ \textbf{\model (Ours)}
& \textbf{71.53} & \textbf{70.68}
& \textbf{53.90} & \textbf{49.20}
& \textbf{59.60}
& \textbf{56.00}
& \textbf{60.15} \\

\midrule

\rowcolor{gray!12}
\multicolumn{9}{c}{\textit{\textbf{Backbone Model: Qwen3-8B}}~\protect\qwenicon} \\
\midrule

\multirow{4}{*}{Base Agent Planning}
& \prompticon\ ReAct                 
& 19.29 & 17.85 & 9.79  & 8.61  & 18.62 & 26.00 & 16.69 \\

& \hspace{0.6em}$\hookrightarrow$ \trainicon\ w/ SFT               
& 67.86 & 57.62 & 57.21 & 50.33 & 56.25 & 42.00 & 55.21 \\

& \prompticon\ Imagine-and-Act                  
& 32.86 & 30.55 & 16.80 & 15.60 & 22.70 & 31.50 & 25.00 \\

& \hspace{0.6em}$\hookrightarrow$ \trainicon\ w/ SFT               
& 68.14 & 72.34 & 58.62 & 42.14 & 46.15 & 46.50 & 55.65 \\

\specialrule{0.25pt}{1pt}{1pt}

\multirow{3}{*}{Test-time Evolving}
& \prompticon\ Reflexion 
& 24.29 & 22.40 & 12.50 & 12.10 & 20.80 & 27.50 & 19.93 \\

& \prompticon\ RAP       
& 28.57 & 26.80 & 15.46 & 27.14 & 12.40 & 28.00 & 23.06 \\

& \prompticon\ ITP$_\text{I}$   
& 41.43 & 39.20 & 19.58 & 19.20 & 24.30 & 35.00 & 29.79 \\

\specialrule{0.25pt}{1pt}{1pt}

\multirow{3}{*}{World Modeling}
& \trainicon\ WKM       
& 79.29 & 76.87 & 60.31 & 47.68 & 61.15 & 40.50 & 60.97 \\

& \trainicon\ IWM       
& 82.14 & 79.50 & 59.27 & 54.30 & 57.30 & 44.00 & 62.75 \\

& \trainicon\ ITP$_\text{R}$   
& 85.71 & 83.20 & 60.20 & 55.30 & 64.75 & 68.00 & 69.53 \\

\specialrule{0.25pt}{1pt}{1pt}

\rowcolor{blue!8}
Co-evolving
& \trainicon\ \textbf{\model (Ours)}
& \textbf{88.57} & \textbf{86.68}
& \textbf{61.85} & \textbf{56.95}
& \textbf{68.10}
& \textbf{70.50}
& \textbf{72.11} \\
\bottomrule
\end{tabular}
}
\caption{
Performance comparison across representative benchmarks, where results on AW, SW, and WS represent task success rate (\%), while results on STB represent solvable pass rate (\%).
Training-free methods are marked with \prompticon, while training-based methods are marked with \trainicon. The best result per backbone model is highlighted in \textbf{bold}.
}
\label{tab:main_results}
\vspace{-6pt}
\end{table*}
\begin{table}[t!]
\centering
\resizebox{\linewidth}{!}{
\begin{tabular}{lccccc}
\toprule
\textbf{Method}
& \textbf{AW}
& \textbf{SW}
& \textbf{WS}
& \textbf{STB}
& \textbf{Avg.} \\
\midrule

\rowcolor{gray!12}
\multicolumn{6}{c}{\textit{API-Called Models}} \\
\midrule

\textit{DeepSeek-V4-Flash}~\protect~\deepseekicon & 80.20 & 77.40 & 75.60 & 61.30 & 73.63 \\
\textit{DeepSeek-V4-Pro}~\protect~\deepseekicon   & 83.35 & 86.80 & 83.10 & 89.80 & 85.76 \\
\textit{GPT-5.4} ~\protect~\chatgpticon          & 91.80 & 92.70 & 84.10 & 87.60 & 89.05 \\
\midrule
\rowcolor{gray!12}
\multicolumn{6}{c}{\textit{Backbone Model: Qwen3-4B}~\protect\qwenicon} \\
\midrule
Vanilla
& 31.25 & 25.00 & 27.60 & 22.90 & 26.69 \\
SFT
& 52.30 & 44.45 & 44.20 & 38.70 & 44.91 \\
On-policy Training
& 78.50 & 81.00 & 70.60 & 73.90 & 76.00 \\
\rowcolor{blue!7}
\textbf{\model (Ours)}
& \textbf{89.20} & \textbf{86.20} & \textbf{86.40} & \textbf{79.80} & \textbf{85.40} \\

\midrule

\rowcolor{gray!12}
\multicolumn{6}{c}{\textit{Backbone Model: Qwen3-8B}~\protect\qwenicon} \\
\midrule
Vanilla
& 38.40 & 30.60 & 33.50 & 28.20 & 32.68 \\
SFT
& 60.05 & 51.90 & 51.70 & 65.40 & 57.26 \\
On-policy Training
& 87.05 & 88.45 & 78.50 & 81.90 & 83.98 \\
\rowcolor{blue!7}
\textbf{\model (Ours)}
& \textbf{93.70} & \textbf{94.15} & \textbf{84.70} & \textbf{88.60} & \textbf{90.29} \\
\bottomrule
\end{tabular}
}
\caption{
Performance comparison across different textual world models.
We report $\Delta$F1 (\%) to measure whether the next predicted world-state captures action-induced state changes after canonicalization.
}
\label{tab:wm_delta_f1_main}
\vspace{-8pt}
\end{table}

\begin{figure*}[t!]
    \centering
    \captionsetup[subfigure]{font=small,skip=2pt}
    \setlength{\abovecaptionskip}{4pt}
    \setlength{\belowcaptionskip}{2pt}

    \begin{subfigure}[t]{0.49\textwidth}
        \centering
        \includegraphics[
            width=\linewidth,
            trim={0pt 0pt 0pt 0pt},
            clip
        ]{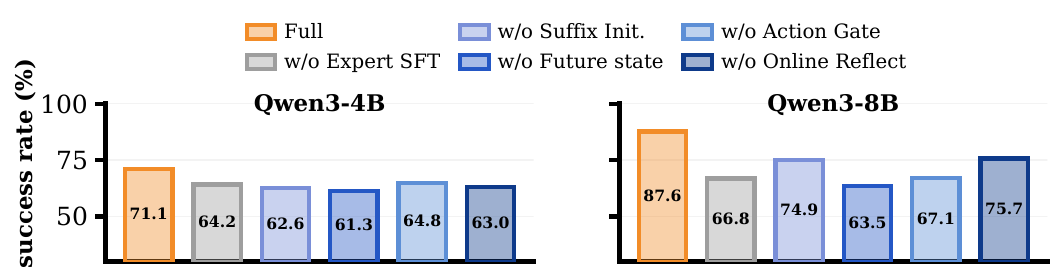}
        \caption{Policy-side ablations.}
        \label{fig:ablation_policy}
    \end{subfigure}
    \hfill
    \begin{subfigure}[t]{0.49\textwidth}
        \centering
        \includegraphics[
            width=\linewidth,
            trim={0pt 0pt 0pt 0pt},
            clip
        ]{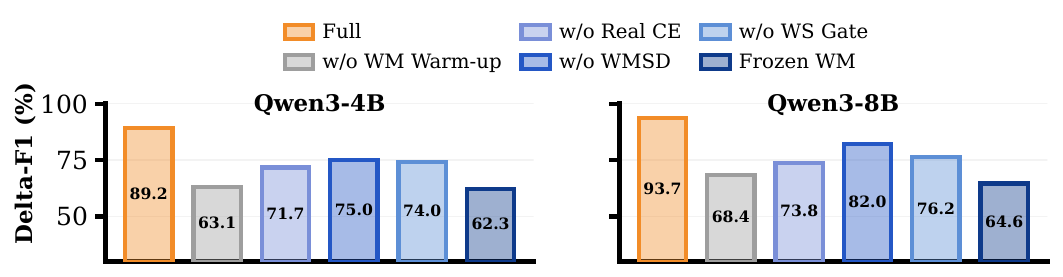}
        \caption{World-model and coupling ablations.}
        \label{fig:ablation_wm}
    \end{subfigure}
    \caption{
    Component ablations of \model on ALFWorld.
    We report leave-one-component-out results on Qwen3-4B and Qwen3-8B.
    Policy-side results represent success rates, while world-model-side results represent $\Delta$F1 scores.
    }
    \label{fig:coevolving_ablation}
    \vspace{-6pt}
\end{figure*}

\paragraph{Benchmarks.}
We employ the following widely used benchmarks for evaluation:
\textbf{ALFWorld (AW)}~\citep{ALFWorld20} instantiates language-conditioned household planning tasks in embodied environments.
\textbf{ScienceWorld (SW)}~\citep{Wang2022ScienceWorld} evaluates multi-step scientific reasoning and experimentation in a textual simulator.
\textbf{WebShop (WS)}~\citep{yao2022webshop} measures how an agent navigates a user in an e-commerce website and assists in purchasing products.
\textbf{StableToolBench (STB)}~\citep{guo-etal-2024-stabletoolbench} evaluates tool-use abilities with a stable virtual API-calling server. 
Detailed descriptions of these benchmarks are provided in Appendix~\ref{app:data}.

\paragraph{Backbone Models.}
We adopt both API-called models and open-source trainable models as backbones.
For API-callled models, we evaluate GPT-5.4~\citep{openai2026gpt54}, DeepSeek-V4-Flash, and DeepSeek-V4-Pro~\citep{deepseek2026v4}, with the same prompting protocol.
For training-based methods, including our co-evolving framework and all trainable baselines, we use Qwen3-4B and Qwen3-8B~\citep{qwen3} as the policy backbones. 
The agent policy and world model employ the same backbone and tokenizer template, while all methods within a backbone group share the same data budget, rollout budget, and evaluation setting.

\paragraph{Baseline Methods.}
We compare \model against the following baselines.
For API-called models, we evaluate GPT-5.4, DeepSeek-V4-Flash, and DeepSeek-V4-Pro under ReAct~\citep{yao2023react} prompting.
We further introduce \emph{Imagine-and-Act}, a prompting-only lookahead baseline that first asks the same model to imagine the one-step consequence of a candidate action and then selects the final action conditioned on the imagined state, without parameter updates or a separately trained world model~\citep{hao-etal-2023-reasoning,liu2026itp}.
For open-source backbone models, we use ReAct and Imagine-and-Act, each with and without supervised fine-tuning (SFT). 
We then compare with test-time evolving methods, including Reflexion~\citep{shinn2023reflexion}, RAP~\citep{hao-etal-2023-reasoning}, and ITP$_\text{I}$~\citep{liu2026itp}.
We also compare with world-modeling baselines, including WKM~\citep{qiao2024agent}, IWM~\citep{zhang2025agent}, and ITP$_\text{R}$~\citep{liu2026itp}.

\paragraph{Evaluation Metrics.} 
To evaluate agent policies, we adopt \textbf{success rate (SR)} as the evaluation metric, defined as the percentage of episodes that successfully complete the given task~\citep{ALFWorld20,wang-etal-2025-steca}. 
For ALFWorld and ScienceWorld, we report SR
on both the seen and unseen test splits. Following~\citet{guo-etal-2024-stabletoolbench}, we report \textbf{solvable pass rate (SoPR)} on the StableToolBench. 
To evaluate the performance of world models, we adopt $\mathbf{\Delta}$\textbf{F1} to measure whether the world model predicts the actual state change caused by the executed action. 
For each action-state transition, we canonicalize states into task-relevant facts and compare only the facts changed by the action, given by:
\[
\begin{aligned}
\Delta \text{F1}=\text{F1}\big(&\Phi(\hat{s}_{t+1}^{\mathrm{wm}})-\Phi(s_t), \Phi(s_{t+1}^{\mathrm{env}})-\Phi(s_t)\big),
\end{aligned}
\]
where $\Phi$ denotes the canonicalization operation.
If the predicted state $\hat{s}_{t+1}^{\mathrm{wm}}$ cannot be parsed into the canonical schema, the resulting $\Delta$F1 is set to 0.

\subsection{Main Results}

\paragraph{Agent Policy Performance.}
Table~\ref{tab:main_results} shows that \model consistently outperforms prior trainable baselines across embodied task planning, web navigation, and tool-use tasks.
On Qwen3-8B, \model improves the average score over the state-of-the-art ITP$_\text{R}$ from 69.53\% to 72.11\%; on Qwen3-4B, the gain is significantly larger, from 51.52\% to 60.15\%.
This suggests that co-evolving is especially helpful for smaller backbones, where future-conditioned reflection can compensate for weaker planning ability.
Compared with prompting-only, test-time evolving, and fixed world-modeling methods, \model further benefits from continuously updating both the policy and the world model.
Notably, Qwen3-8B with \model surpasses GPT-5.4 with ReAct and DeepSeek-V4-Pro with ReAct, and approaches DeepSeek-V4-Pro with Imagine-and-Act.

\paragraph{World Model Performance.}
Table~\ref{tab:wm_delta_f1_main} shows that \model also improves world-model prediction quality, achieving  85.40\% and 90.29\% average $\Delta$F1 on Qwen3-4B and Qwen3-8B, respectively.
Notably, Qwen3-8B with \model even slightly surpasses GPT-5.4 in $\Delta$F1, while Qwen3-4B with \model approaches DeepSeek-V4-Pro and clearly outperforms DeepSeek-V4-Flash.
Compared with vanilla prompting, SFT, and on-policy training baselines, the consistent gains achieved by \model indicate that real-transition supervision alone is insufficient, and that self-distilled on-policy adaptation is crucial for modeling policy-induced state changes.
These results show that co-evolving improves the capabilities of world models.

\subsection{Ablation Study}
To evaluate the contribution of individual components within \model, we conduct comprehensive ablations on the ALFWorld benchmark using both Qwen3-4B and Qwen3-8B as backbone models.

\paragraph{Policy-Side Ablations.}  
As shown in Figure~\ref{fig:coevolving_ablation}(a), the full \model framework achieves the highest task success rate across both model scales. 
Removing any module leads to a substantial performance decline. Notably, omitting the future-state input (\textit{w/o Future State}) causes the most severe drop, reducing the success rate by 9.8\% and 24.1\% on the 4B and 8B models, respectively. 
Furthermore, removing the initial expert supervised fine-tuning (\textit{w/o Expert SFT}) or the reflection-mode initialization also heavily impairs performance, confirming the necessity of high-quality trajectory priors and robust action-refinement training. 
The absence of the action gate and online reflection similarly degrades the success rate, proving that adaptive reliance on world models and continuous policy refinement are indispensable.

\paragraph{World-model and Coupling Ablations.} 
To assess the world model's state-tracking fidelity, we report the $\Delta$F1 score in Figure~\ref{fig:coevolving_ablation}(b). 
The full framework significantly outperforms all variants. 
Most importantly, freezing the world model after initial training (\textit{Frozen WM}) results in the most drastic deterioration in $\Delta$F1, dropping by a massive 26.9\% and 29.1\% on the 4B and 8B models. This strongly validates our core hypothesis: static world models quickly become misaligned during long-horizon tasks, and continuous, dynamic co-evolution is essential for maintaining accurate environment modeling. 
Additionally, eliminating the world model warm-up phase (\textit{w/o WM Warm-up}) severely hinders the model's baseline predictive ability. Other coupling mechanisms, including real-state supervision (\textit{w/o Real CE}), world-model self-distillation (\textit{w/o WMSD}), and the world-state gate (\textit{w/o WS Gate}), all contribute positively to the final prediction accuracy, ensuring the world model remains robust and closely aligned with the real environment dynamics.

\subsection{In-depth Analyses and Discussions}

\begin{figure}[t]
    \centering
    \begin{subfigure}[t]{\linewidth}
        \centering
        \includegraphics[width=\linewidth]{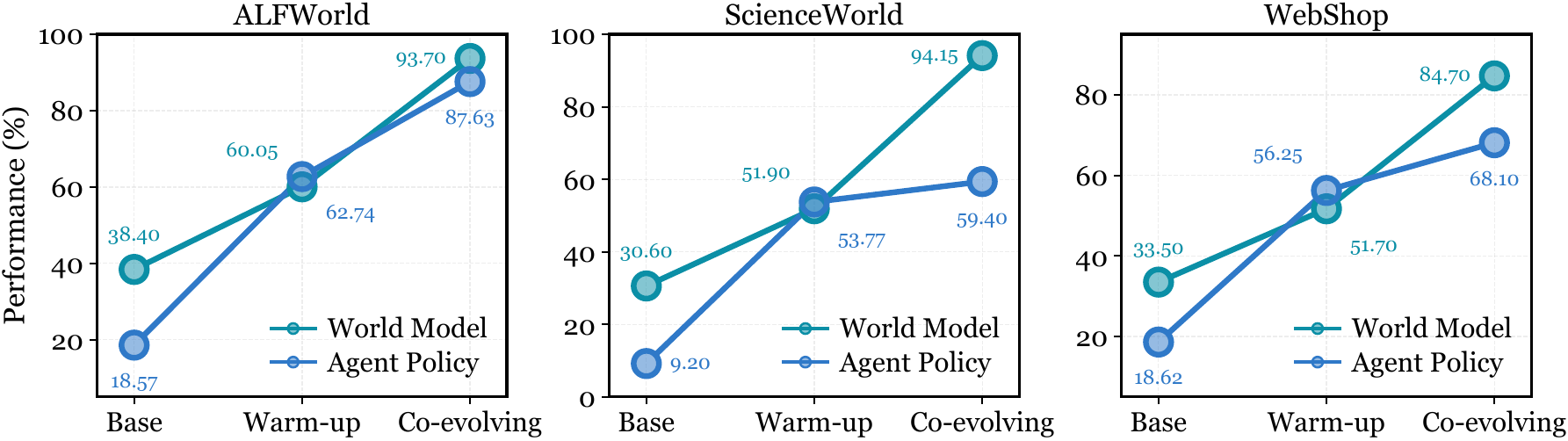}
        \caption{Stage-wise performance comparison.}
        \label{fig:stage_comparison}
    \end{subfigure}
    \vspace{0.5em}
    \begin{subfigure}[t]{\linewidth}
        \centering
        \includegraphics[width=\linewidth]{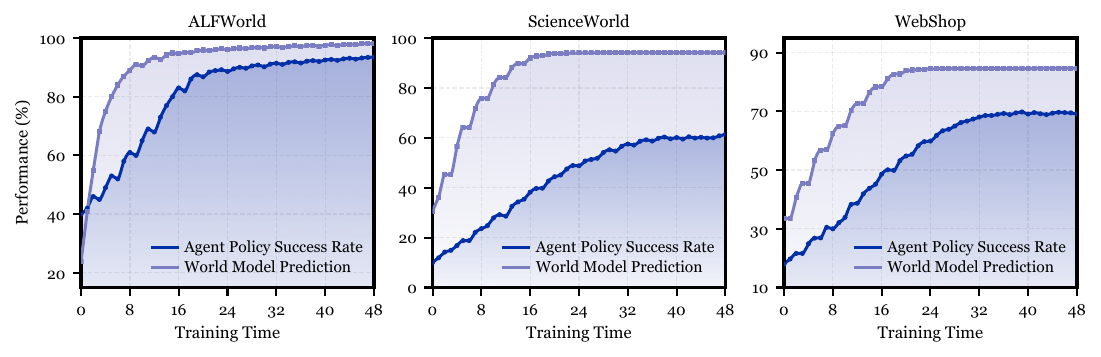}
        \caption{Learning dynamics during co-evolving training.}
        \label{fig:learning_dynamics}
    \end{subfigure}
    \vspace{-6pt}
    \caption{
    Learning dynamics of co-evolving.
    The upper panel compares the world model and agent policy across base, warm-up, and co-evolving stages.
    The lower panel shows that both the world model and agent policy stably improve during co-evolving training.
    }
    \label{fig:co_evolving_performance_analysis}
    \vspace{-6pt}
\end{figure}

\paragraph{Learning Dynamics of Co-Evolving.}
We analyze the training dynamics during the co-evolving process.
As shown in Figure~\ref{fig:stage_comparison}, both the world model and the agent policy improve consistently from the base model to warm-up and co-evolving stages, indicating that supervised warm-up provides a useful initialization and online co-evolving further strengthens both components.
Figure~\ref{fig:learning_dynamics} shows that the world model reaches a relatively stable prediction accuracy after early training, while the agent policy continues to improve with small fluctuations.
This trend suggests that a stabilized world model can provide increasingly useful future-state signals for policy reflection, and the policy can further benefit from high-quality online refinement during co-evolving training.

\begin{table}[t!]
\centering
\resizebox{0.85\linewidth}{!}{
\setlength{\tabcolsep}{8pt}
\renewcommand{\arraystretch}{1.15}
\begin{tabular}{lccc}
\toprule
\textbf{Benchmark} 
& \textbf{URR ($\downarrow$) } 
& \textbf{HRR ($\downarrow$) } 
& \textbf{BRP ($\uparrow$) } \\
\midrule
ALFWorld     & 0.16 & 0.08 & 0.92 \\
WebShop      & 0.07 & 0.09 & 0.87 \\
ScienceWorld &  0.12& 0.11 & 0.93 \\
\bottomrule
\end{tabular}}
\caption{Effect of the future-aware reflection stage.}
\label{tab:revise_stage}
\vspace{-6pt}
\end{table}

\paragraph{Effect of Reflection.}
We evaluate the future-aware reflection stage using three diagnostic metrics: unnecessary revision rate \textbf{(URR)}, harmful revision rate \textbf{(HRR)}, and beneficial revision precision \textbf{(BRP)}. 
URR measures over-intervention on already successful draft actions, while HRR measures destructive revisions that turn successful drafts into failed reflected actions. BRP measures the precision of useful revisions. All metrics lie in the $[0,1]$ range. 
As reported in Table~\ref{tab:revise_stage}, the consistently low URR/HRR and high BRP indicate that reflection contributes critical refinement while avoiding unnecessary or harmful revisions. 
We provide more detailed discussions in Appendix~\ref{app:mechanism_clarification}.

\begin{figure}[t]
    \centering
    \includegraphics[width=0.98\linewidth]{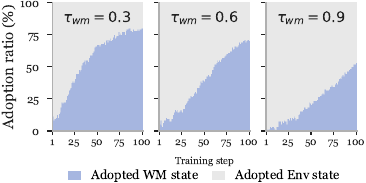}
    \caption{Adoption ratio of the world state gate (Qwen3-8B as the backbone model on ALFWorld).}
    \label{fig:wm-gate}
    \vspace{-6pt}
\end{figure}

\paragraph{Effect of World State Gate.}

We examine whether the world-state gate could adaptively select predicted world-model states during co-evolving training.
As shown in Figure~\ref{fig:wm-gate}, the world model state adoption ratio consistently increases along training under different thresholds, suggesting that the world model produces increasingly reliable future-state predictions.
A lower threshold $\tau_{wm}=0.3$ enables earlier and more frequent adoption, whereas a higher threshold $\tau_{wm}=0.9$ keeps the gate conservative.
The middle setting $\tau_{wm}=0.6$ provides a balanced curriculum, suppressing noisy early predictions while gradually exposing the policy to useful world model-guided states.
We therefore use $\tau_{wm}=0.6$ in the main experiments.

\begin{figure}[t] 
\centering 
\includegraphics[width=0.96\linewidth]{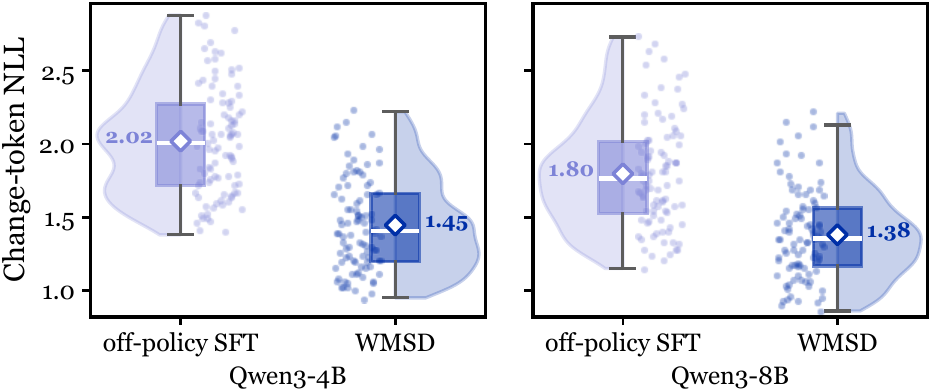} 
\caption{Change-token NLL under off-policy SFT and WMSD (Qwen3-4B and Qwen3-8B on ALFWorld). } 
\label{fig:wmsd_change_token_nll} 
\vspace{-6pt}
\end{figure}

\paragraph{Analysis of On-Policy Self-Distillation for World Models.}
To investigate how on-policy self-distillation facilitates world-model evolution, we contrast it with standard off-policy SFT. 
While off-policy SFT relies on a single hard next-state target, world-model self-distillation (WMSD) couples the real-state anchor with token-level soft guidance from the teacher model. 
This mechanism provides denser, more informative supervision over plausible transition tokens. 
We isolate and evaluate this effect by measuring the prediction loss on \emph{action-induced change tokens}. 
As illustrated in Figure~\ref{fig:wmsd_change_token_nll}, WMSD consistently yields a lower negative log-likelihood (NLL). 
Specifically, WMSD reduces the mean change-token NLL from 2.02 to 1.45 and from 1.80 to 1.38, respectively. 
These results demonstrate that rather than merely memorizing static background text, WMSD effectively trains the world model to capture transition-critical state changes with greater accuracy and consistency.

\section{Related Work}
\label{sec:related_work}

\paragraph{World Models for Planning.}
World models have long been viewed as a key mechanism for autonomous intelligence and model-based reinforcement learning, enabling agents to predict future states and evaluate action consequences before execution~\citep{lecun2022path,hafner2023mastering}. With the rise of LLM agents, this idea has been extended to \emph{textual world models}, where language models predict environment feedback, state transitions, and action outcomes for task planning~\citep{guan2023leveraging,hao-etal-2023-reasoning,Gu2025WebDreamer,Cheng2025WebATLAS}. These models provide an explicit foresight perspective, allowing the agent policy to obtain additional future-state information and thus improve reasoning and planning beyond the current observation alone~\citep{xiang2024language,sun2024enhancing,wang-etal-2025-world,liu2026itp}. Recent work further explores world models for agent improvement, such as WebEvolver, which leverages a co-evolving world model for web-agent self-improvement~\citep{fang-etal-2025-webevolver}. In contrast to prior methods that mainly use world models as planning components, our work studies a closed-loop formulation where the textual world model and the agent policy are jointly optimized and mutually reinforced through interaction.

\paragraph{LLM Agent Evolving.}
Initial efforts in agent evolution focused on self-reflection~\citep{shinn2023reflexion} or retrieving past experiences~\citep{wei2012coaching,majumder2023clin,qian2024ice}. 
Moving beyond prompting, parametric methods optimize policies by fine-tuning on expert trajectories, correcting execution errors, and calibrating behaviors~\citep{chen2023fireact,song-etal-2024-trial,wang-etal-2024-e2cl,wang-etal-2025-steca}. 
To achieve autonomous self-improvement~\citep{song2026survey}, recent paradigms emphasize continuous learning mechanisms. These methods enable agents to dynamically refine policies via environmental feedback~\citep{xi2024agentgym,sheng2026rlcer}, abstract historical experiences into symbolic rules~\citep{zhou2024symbolic,lu2023self}, or acquire new capabilities through open-ended skill-bank curation~\citep{xu2026ael,wu2026cosplay}. 
While some advanced methods incorporate world models to facilitate adaptive lookahead planning~\citep{wang2025vagen,liu2026itp}, they typically treat the models as isolated, auxiliary simulators. 
In contrast, our \model uniquely co-evolves both world models and agent policies through continuous closed-loop interactions.

\section{Conclusion}

In this paper, we introduced \model, a closed-loop framework that co-evolves textual world models and agent policies through interaction. 
By combining on-policy self-distillation with future-aware reflection, \model continually adapts the world model to the agent's evolving interaction distribution and uses its predictions to refine agent decision-making. 
Extensive experiments across embodied task planning, web navigation, and tool-use benchmarks demonstrate consistent improvements in both task success and world-model prediction accuracy. 
These findings highlight the potential of world models to move beyond static planning components and become evolving sources of future-aware supervision, enabling LLM agents to improve through their own interaction experience without relying on external rewards or verifiers.

\section*{Limitations}
While \model shows consistent gains across multiple agent benchmarks, two limitations remain. 
First, \model assumes that environment states and action outcomes can be represented as text. This fits web navigation, tool-use tasks, and text-based environments, but may not fully cover multimodal settings within physical environments. 
Second, \model requires an additional world-model call at inference time before future-aware reflection. While this cost is moderate due to one-step imagination rather than multi-branch search or long-horizon rollouts, it may affect latency-sensitive applications. 
Future work can reduce this overhead through selective imagination, lightweight world model caching, or asynchronous prediction.

\section*{Ethics Statement}
We follow the ethical guidelines for the academic use of LLMs. Our experiments are conducted on simulated agent benchmarks and do not involve human subjects or personally identifiable information. 
We cite and comply with the licenses of the models, datasets, and software used in this work. 
We acknowledge that world-model-based lookahead may introduce risks if deployed beyond controlled environments, since inaccurate future-state predictions could lead to unreliable actions in open-world tool-use or robotic systems. 
Therefore, \model should be used with appropriate safety checks in real-world or high-stakes scenarios. 
AI assistants were partially used for coding and language polishing, while all core ideas, experiments, and findings are the original work of the authors.

\section*{Acknowledgements}
This work was supported by the Fundamental Research Funds for the Central Universities (YJ202620 and 2026SCUQJTX022B), the General Research Fund (GRF) of the Research Grants Council of Hong Kong (PolyU 15205325), and was supported in part by the PolyU Postdoc Matching Fund (4-W40Z).
The authors would like to thank the anonymous reviewers for their valuable feedback and constructive suggestions.

\bibliography{custom}

@inproceedings{xiang2024language,
  title={Language Models Meet World Models: Embodied Experiences Enhance Language Models},
  author={Xiang, Jiannan and Tao, Tianhua and Gu, Yi and Shu, Tianmin and Wang, Zirui and Yang, Zichao and Hu, Zhiting},
  booktitle={Advances in Neural Information Processing Systems},
  volume={36},
  pages={75392--75412},
  publisher={Curran Associates, Inc.},
  doi={10.52202/075280-3295},
  url={https://proceedings.neurips.cc/paper_files/paper/2023/hash/ee6630dcbcff857026e474fc857aa9f0-Abstract-Conference.html},
  year={2023}
}

@inproceedings{Wang2022ScienceWorld,
    title = "{S}cience{W}orld: Is your Agent Smarter than a 5th Grader?",
    author = "Wang, Ruoyao  and
      Jansen, Peter  and
      C{\^o}t{\'e}, Marc-Alexandre  and
      Ammanabrolu, Prithviraj",
    booktitle = "Proceedings of the 2022 Conference on Empirical Methods in Natural Language Processing",
    month = dec,
    year = "2022",
    address = "Abu Dhabi, United Arab Emirates",
    publisher = "Association for Computational Linguistics",
    pages = "11279--11298",
    doi = "10.18653/v1/2022.emnlp-main.775",
    url = "https://aclanthology.org/2022.emnlp-main.775/",
}

@article{qwen3,
    title={{Qwen3} Technical Report}, 
    author = {
  Yang, An and
  Li, Anfeng and
  Yang, Baosong and
  Zhang, Beichen and
  Hui, Binyuan and
  Zheng, Bo and
  Yu, Bowen and
  Gao, Chang and
  Huang, Chengen and
  Lv, Chenxu and
  Zheng, Chujie and
  Liu, Dayiheng and
  Zhou, Fan and
  Huang, Fei and
  Hu, Feng and
  Ge, Hao and
  Wei, Haoran and
  Lin, Huan and
  Tang, Jialong and
  Yang, Jian and
  Tu, Jianhong and
  Zhang, Jianwei and
  Yang, Jianxin and
  Yang, Jiaxi and
  Zhou, Jing and
  Zhou, Jingren and
  Lin, Junyang and
  Dang, Kai and
  Bao, Keqin and
  Yang, Kexin and
  Yu, Le and
  Deng, Lianghao and
  Li, Mei and
  Xue, Mingfeng and
  Li, Mingze and
  Zhang, Pei and
  Wang, Peng and
  Zhu, Qin and
  Men, Rui and
  Gao, Ruize and
  Liu, Shixuan and
  Luo, Shuang and
  Li, Tianhao and
  Tang, Tianyi and
  Yin, Wenbiao and
  Ren, Xingzhang and
  Wang, Xinyu and
  Zhang, Xinyu and
  Ren, Xuancheng and
  Fan, Yang and
  Su, Yang and
  Zhang, Yichang and
  Zhang, Yinger and
  Wan, Yu and
  Liu, Yuqiong and
  Wang, Zekun and
  Cui, Zeyu and
  Zhang, Zhenru and
  Zhou, Zhipeng and
  Qiu, Zihan},
    journal = {arXiv preprint arXiv:2505.09388},
    year={2025}
}

@inproceedings{yao2023react,
  title = {{ReAct}: Synergizing Reasoning and Acting in Language Models},
  author = {Yao, Shunyu and Zhao, Jeffrey and Yu, Dian and Du, Nan and Shafran, Izhak and Narasimhan, Karthik and Cao, Yuan},
  booktitle = {The Eleventh International Conference on Learning Representations},
  year = {2023},
  url = {https://openreview.net/forum?id=WE_vluYUL-X},
}

@inproceedings{wang-etal-2025-steca,
    title = "{ST}e{C}a: Step-level Trajectory Calibration for {LLM} Agent Learning",
    author = "Wang, Hanlin  and
      Wang, Jian  and
      Leong, Chak Tou  and
      Li, Wenjie",
    editor = "Che, Wanxiang  and
      Nabende, Joyce  and
      Shutova, Ekaterina  and
      Pilehvar, Mohammad Taher",
    booktitle = "Findings of the Association for Computational Linguistics: ACL 2025",
    month = jul,
    year = "2025",
    address = "Vienna, Austria",
    publisher = "Association for Computational Linguistics",
    url = "https://aclanthology.org/2025.findings-acl.604/",
    doi = "10.18653/v1/2025.findings-acl.604",
    pages = "11597--11614",
}

@article{wei2012coaching,
  title={Coaching the Exploration and Exploitation in Active Learning for Interactive Video Retrieval},
  author={Wei, Xiao-Yong and Yang, Zhen-Qun},
  journal={IEEE Transactions on Image Processing},
  volume={22},
  number={3},
  pages={955--968},
  year={2013},
  publisher={IEEE},
  doi={10.1109/TIP.2012.2222902},
  url={https://doi.org/10.1109/TIP.2012.2222902}
}

@inproceedings{song-etal-2024-trial,
    title = "Trial and Error: Exploration-Based Trajectory Optimization of {LLM} Agents",
    author = "Song, Yifan  and
      Yin, Da  and
      Yue, Xiang  and
      Huang, Jie  and
      Li, Sujian  and
      Lin, Bill Yuchen",
    editor = "Ku, Lun-Wei  and
      Martins, Andre  and
      Srikumar, Vivek",
    booktitle = "Proceedings of the 62nd Annual Meeting of the Association for Computational Linguistics (Volume 1: Long Papers)",
    month = aug,
    year = "2024",
    address = "Bangkok, Thailand",
    publisher = "Association for Computational Linguistics",
    url = "https://aclanthology.org/2024.acl-long.409/",
    doi = "10.18653/v1/2024.acl-long.409",
    pages = "7584--7600"
}

@inproceedings{shinn2023reflexion,
  title={Reflexion: Language agents with verbal reinforcement learning},
  author={Shinn, Noah and Cassano, Federico and Gopinath, Ashwin and Narasimhan, Karthik and Yao, Shunyu},
  booktitle={Advances in Neural Information Processing Systems},
  volume={36},
  pages={8634--8652},
  year={2023}
}

@article{hafner2023mastering,
  title={Mastering diverse domains through world models},
  author={Hafner, Danijar and Pasukonis, Jurgis and Ba, Jimmy and Lillicrap, Timothy},
  journal={arXiv preprint arXiv:2301.04104},
  year={2023}
}

@inproceedings{wang-etal-2025-world,
    title = "World Modeling Makes a Better Planner: Dual Preference Optimization for Embodied Task Planning",
    author = "Wang, Siyin  and
      Fei, Zhaoye  and
      Cheng, Qinyuan  and
      Zhang, Shiduo  and
      Cai, Panpan  and
      Fu, Jinlan  and
      Qiu, Xipeng",
    editor = "Che, Wanxiang  and
      Nabende, Joyce  and
      Shutova, Ekaterina  and
      Pilehvar, Mohammad Taher",
    booktitle = "Proceedings of the 63rd Annual Meeting of the Association for Computational Linguistics (Volume 1: Long Papers)",
    month = jul,
    year = "2025",
    address = "Vienna, Austria",
    publisher = "Association for Computational Linguistics",
    url = "https://aclanthology.org/2025.acl-long.1044/",
    doi = "10.18653/v1/2025.acl-long.1044",
    pages = "21518--21537",
}

@inproceedings{hao-etal-2023-reasoning,
    title = "Reasoning with Language Model is Planning with World Model",
    author = "Hao, Shibo  and
      Gu, Yi  and
      Ma, Haodi  and
      Hong, Joshua  and
      Wang, Zhen  and
      Wang, Daisy  and
      Hu, Zhiting",
    editor = "Bouamor, Houda  and
      Pino, Juan  and
      Bali, Kalika",
    booktitle = "Proceedings of the 2023 Conference on Empirical Methods in Natural Language Processing",
    month = dec,
    year = "2023",
    address = "Singapore",
    publisher = "Association for Computational Linguistics",
    url = "https://aclanthology.org/2023.emnlp-main.507/",
    doi = "10.18653/v1/2023.emnlp-main.507",
    pages = "8154--8173",
}

@misc{lecun2022path,
  title={A Path Towards Autonomous Machine Intelligence},
  author={LeCun, Yann},
  year={2022},
  howpublished={OpenReview},
  note={Version 0.9.2, 2022-06-27},
  url={https://openreview.net/forum?id=BZ5a1r-kVsf}
}

@inproceedings{fang-etal-2025-webevolver,
    title = "{W}eb{E}volver: Enhancing Web Agent Self-Improvement with Co-evolving World Model",
    author = "Fang, Tianqing  and
      Zhang, Hongming  and
      Zhang, Zhisong  and
      Ma, Kaixin  and
      Yu, Wenhao  and
      Mi, Haitao  and
      Yu, Dong",
    editor = "Christodoulopoulos, Christos  and
      Chakraborty, Tanmoy  and
      Rose, Carolyn  and
      Peng, Violet",
    booktitle = "Proceedings of the 2025 Conference on Empirical Methods in Natural Language Processing",
    month = nov,
    year = "2025",
    address = "Suzhou, China",
    publisher = "Association for Computational Linguistics",
    url = "https://aclanthology.org/2025.emnlp-main.454/",
    doi = "10.18653/v1/2025.emnlp-main.454",
    pages = "8959--8975",
    ISBN = "979-8-89176-332-6"
}

@inproceedings{zhang2025agent,
  title     = {Agent Learning via Early Experience},
  author = {
  Zhang, Kai and
  Chen, Xiangchao and
  Liu, Bo and
  Xue, Tianci and
  Liao, Zeyi and
  Liu, Zhihan and
  Wang, Xiyao and
  Ning, Yuting and
  Chen, Zhaorun and
  Fu, Xiaohan and
  Xie, Jian and
  Sun, Yuxuan and
  Gou, Boyu and
  Qi, Qi and
  Meng, Zihang and
  Yang, Jianwei and
  Zhang, Ning and
  Li, Xian and
  Shah, Ashish and
  Huynh, Dat and
  Li, Hengduo and
  Yang, Zi and
  Cao, Xuefei and
  Jang, Lawrence Keunho and
  Zhou, Shuyan and
  Zhu, Jiacheng and
  Sun, Huan and
  Weston, Jason E and
  Su, Yu and
  Wu, Yifan},
  booktitle = {Proceedings of the 43rd International Conference on Machine Learning},
  series    = {Proceedings of Machine Learning Research},
  volume    = {306},
  year      = {2026},
  address   = {Seoul, South Korea},
  publisher = {PMLR},
  url       = {https://openreview.net/pdf/b090aea758a96038a0c9b91036f8cf16813fb4bc.pdf}
}

@inproceedings{qiao2024agent,
  title={Agent planning with world knowledge model},
  author={Qiao, Shuofei and Fang, Runnan and Zhang, Ningyu and Zhu, Yuqi and Chen, Xiang and Deng, Shumin and Jiang, Yong and Xie, Pengjun and Huang, Fei and Chen, Huajun},
  booktitle={Advances in Neural Information Processing Systems},
  volume={37},
  pages={114843--114871},
  year={2024}
}

@inproceedings{wang-etal-2024-e2cl,
    title = "{E}$^2${CL}: Exploration-based Error Correction Learning for Embodied Agents",
    author = "Wang, Hanlin  and
      Leong, Chak Tou  and
      Wang, Jian  and
      Li, Wenjie",
    editor = "Al-Onaizan, Yaser  and
      Bansal, Mohit  and
      Chen, Yun-Nung",
    booktitle = "Findings of the Association for Computational Linguistics: EMNLP 2024",
    month = nov,
    year = "2024",
    address = "Miami, Florida, USA",
    publisher = "Association for Computational Linguistics",
    url = "https://aclanthology.org/2024.findings-emnlp.448/",
    doi = "10.18653/v1/2024.findings-emnlp.448",
    pages = "7626--7639",
}

@inproceedings{guan2023leveraging,
  title={Leveraging pre-trained large language models to construct and utilize world models for model-based task planning},
  author={Guan, Lin and Valmeekam, Karthik and Sreedharan, Sarath and Kambhampati, Subbarao},
  booktitle={Advances in Neural Information Processing Systems},
  volume={36},
  pages={79081--79094},
  year={2023}
}

@inproceedings{wang2025vagen,
  title={{VAGEN}: Reinforcing World Model Reasoning for Multi-Turn {VLM} Agents},
  author={Wang, Kangrui and Zhang, Pingyue and Wang, Zihan and Gao, Yaning and Li, Linjie and Wang, Qineng and Chen, Hanyang and Lu, Yiping and Yang, Zhengyuan and Wang, Lijuan and Krishna, Ranjay and Wu, Jiajun and Li, Fei-Fei and Choi, Yejin and Li, Manling},
  booktitle={Advances in Neural Information Processing Systems},
  volume={38},
  pages={172871--172933},
  publisher={Curran Associates, Inc.},
  doi={10.52202/085713-5754},
  url={https://proceedings.neurips.cc/paper_files/paper/2025/hash/fc6688d75adde86b9df910769c1d02e3-Abstract-Conference.html},
  year={2025}
}

@inproceedings{li2025word,
  title     = {From Word to World: Can Large Language Models be Implicit Text-based World Models?},
  author    = {Li, Yixia and Wang, Hongru and Qiu, Jiahao and Yin, Zhenfei and Zhang, Dongdong and Qian, Cheng and Li, Zeping and Ma, Xiaoteng and Chen, Guanhua and Ji, Heng},
  editor    = {Liakata, Maria and Moreira, Viviane P. and Zhang, Jiajun and Jurgens, David},
  booktitle = {Proceedings of the 64th Annual Meeting of the Association for Computational Linguistics (Volume 1: Long Papers)},
  month     = jul,
  year      = {2026},
  address   = {San Diego, California, United States},
  publisher = {Association for Computational Linguistics},
  pages     = {8084--8111},
  doi       = {10.18653/v1/2026.acl-long.366},
  url       = {https://aclanthology.org/2026.acl-long.366/},
  isbn      = {979-8-89176-390-6}
}

@inproceedings{yao2022webshop,
  author    = {Shunyu Yao and Howard Chen and John Yang and Karthik Narasimhan},
  title     = {{WebShop}: Towards Scalable Real-World Web Interaction with Grounded Language Agents},
  booktitle = {Advances in Neural Information Processing Systems},
  volume    = {35},
  pages     = {20744--20757},
  year      = {2022}
}

@article{chen2023fireact,
  title={{FireAct}: Toward Language Agent Fine-Tuning},
  author={Chen, Baian and Shu, Chang and Shareghi, Ehsan and Collier, Nigel and Narasimhan, Karthik and Yao, Shunyu},
  journal={arXiv preprint arXiv:2310.05915},
  year={2023}
}

@inproceedings{guo-etal-2024-stabletoolbench,
    title = "{S}table{T}ool{B}ench: Towards Stable Large-Scale Benchmarking on Tool Learning of Large Language Models",
    author = "Guo, Zhicheng and
      Cheng, Sijie and
      Wang, Hao and
      Liang, Shihao and
      Qin, Yujia and
      Li, Peng and
      Liu, Zhiyuan and
      Sun, Maosong and
      Liu, Yang",
    editor = "Ku, Lun-Wei and
      Martins, Andre and
      Srikumar, Vivek",
    booktitle = "Findings of the Association for Computational Linguistics: ACL 2024",
    month = aug,
    year = "2024",
    address = "Bangkok, Thailand",
    publisher = "Association for Computational Linguistics",
    url = "https://aclanthology.org/2024.findings-acl.664/",
    doi = "10.18653/v1/2024.findings-acl.664",
    pages = "11143--11156"
}

@article{liu2026itp,
  title        = {Imagine-then-Plan: Agent Learning from Adaptive Lookahead with World Models},
  author       = {Liu, Youwei and Wang, Jian and Wang, Hanlin and Guo, Beichen and Li, Wenjie},
  journal      = {arXiv preprint arXiv:2601.08955},
  year         = {2026},
  doi          = {10.48550/arXiv.2601.08955},
  url          = {https://arxiv.org/abs/2601.08955}
}

@article{wang2024survey,
  title={A Survey on Large Language Model Based Autonomous Agents},
  author = {
  Wang, Lei and
  Ma, Chen and
  Feng, Xueyang and
  Zhang, Zeyu and
  Yang, Hao and
  Zhang, Jingsen and
  Chen, Zhiyuan and
  Tang, Jiakai and
  Chen, Xu and
  Lin, Yankai and
  Zhao, Wayne Xin and
  Wei, Zhewei and
  Wen, Jirong},
  journal={Frontiers of Computer Science},
  volume={18},
  number={6},
  pages={186345},
  year={2024},
  publisher={Springer},
  doi={10.1007/s11704-024-40231-1},
  url={https://doi.org/10.1007/s11704-024-40231-1}
}

@inproceedings{li2024embodied,
  title={Embodied Agent Interface: Benchmarking {LLMs} for Embodied Decision Making},
  author={Li, Manling and Zhao, Shiyu and Wang, Qineng and Wang, Kangrui and Zhou, Yu and Srivastava, Sanjana and Gokmen, Cem and Lee, Tony and Li, Li Erran and Zhang, Ruohan and Liu, Weiyu and Liang, Percy and Fei-Fei, Li and Mao, Jiayuan and Wu, Jiajun},
  booktitle={Advances in Neural Information Processing Systems},
  volume={37},
  pages={100428--100534},
  publisher={Curran Associates, Inc.},
  doi={10.52202/079017-3188},
  url={https://proceedings.neurips.cc/paper_files/paper/2024/hash/b631da756d1573c24c9ba9c702fde5a9-Abstract-Datasets_and_Benchmarks_Track.html},
  year={2024}
}

@inproceedings{chae2025web,
  title={Web Agents with World Models: Learning and Leveraging Environment Dynamics in Web Navigation},
  author={Chae, Hyungjoo and Kim, Namyoung and Ong, Kai and Gwak, Minju and Song, Gwanwoo and Kim, Jihoon and Kim, Sunghwan and Lee, Dongha and Yeo, Jinyoung},
  booktitle={International Conference on Learning Representations},
  volume={2025},
  pages={63707--63738},
  url={https://proceedings.iclr.cc/paper_files/paper/2025/hash/a00548031e4647b13042c97c922fadf1-Abstract-Conference.html},
  year={2025}
}

@inproceedings{sun2024enhancing,
  title={Enhancing Agent Learning through World Dynamics Modeling},
  author={Sun, Zhiyuan and Shi, Haochen and C{\^o}t{\'e}, Marc-Alexandre and Berseth, Glen and Yuan, Xingdi and Liu, Bang},
  booktitle={Findings of the Association for Computational Linguistics: EMNLP 2024},
  pages={3534--3568},
  month={nov},
  address={Miami, Florida, USA},
  publisher={Association for Computational Linguistics},
  doi={10.18653/v1/2024.findings-emnlp.202},
  url={https://aclanthology.org/2024.findings-emnlp.202/},
  year={2024}
}

@misc{openai2026gpt54,
  title        = {Introducing {GPT-5.4}},
  author       = {{OpenAI}},
  year         = {2026},
  month        = mar,
  day          = {5},
  howpublished = {OpenAI},
  url          = {https://openai.com/index/introducing-gpt-5-4/},
  note         = {Accessed: 2026-05-24}
}

@article{song2026survey,
  title={A Survey of On-Policy Distillation for Large Language Models},
  author={Song, Mingyang and Zheng, Mao},
  journal={arXiv preprint arXiv:2604.00626},
  year={2026}
}

@article{sheng2026rlcer,
  title   = {Reinforcing Chain-of-Thought Reasoning with Self-Evolving Rubrics},
  author  = {Sheng, Leheng and Ma, Wenchang and Hong, Ruixin and Wang, Xiang and Zhang, An and Chua, Tat-Seng},
  journal = {arXiv preprint arXiv:2602.10885},
  year    = {2026}
}

@misc{deepseek2026v4,
  title        = {{DeepSeek-V4} Preview Release},
  author       = {{DeepSeek-AI}},
  year         = {2026},
  howpublished = {DeepSeek API Docs},
  url          = {https://api-docs.deepseek.com/news/news260424},
  note         = {Accessed: 2026-05-25}
}

@inproceedings{majumder2023clin,
  title     = {{CLIN}: A Continually Learning Language Agent for Rapid Task Adaptation and Generalization},
  author    = {Majumder, Bodhisattwa Prasad and Mishra, Bhavana Dalvi and Jansen, Peter and Tafjord, Oyvind and Tandon, Niket and Zhang, Li and Callison-Burch, Chris and Clark, Peter},
  booktitle = {First Conference on Language Modeling},
  year      = {2024},
  url       = {https://openreview.net/forum?id=xS6zx1aBI9}
}

@misc{qian2024ice,
  title        = {Investigate-Consolidate-Exploit: A General Strategy for Inter-Task Agent Self-Evolution},
  author       = {Qian, Cheng and Liang, Shihao and Qin, Yujia and Ye, Yining and Cong, Xin and Lin, Yankai and Wu, Yesai and Liu, Zhiyuan and Sun, Maosong},
  year         = {2024},
  eprint       = {2401.13996},
  archivePrefix = {arXiv},
  primaryClass = {cs.CL},
  doi          = {10.48550/arXiv.2401.13996}
}

@inproceedings{xi2024agentgym,
  title     = {{AgentGym}: Evaluating and Training Large Language Model-based Agents across Diverse Environments},
  author    = {Xi, Zhiheng and Ding, Yiwen and Chen, Wenxiang and Hong, Boyang and Guo, Honglin and Wang, Junzhe and Guo, Xin and Yang, Dingwen and Liao, Chenyang and He, Wei and Gao, Songyang and Chen, Lu and Zheng, Rui and Zou, Yicheng and Gui, Tao and Zhang, Qi and Qiu, Xipeng and Huang, Xuanjing and Wu, Zuxuan and Jiang, Yu-Gang},
  editor    = {Che, Wanxiang and Nabende, Joyce and Shutova, Ekaterina and Pilehvar, Mohammad Taher},
  booktitle = {Proceedings of the 63rd Annual Meeting of the Association for Computational Linguistics (Volume 1: Long Papers)},
  month     = jul,
  year      = {2025},
  address   = {Vienna, Austria},
  publisher = {Association for Computational Linguistics},
  pages     = {27914--27961},
  doi       = {10.18653/v1/2025.acl-long.1355},
  url       = {https://aclanthology.org/2025.acl-long.1355/},
  isbn      = {979-8-89176-251-0}
}

@misc{lu2023self,
  title        = {{SELF}: Self-Evolution with Language Feedback},
  author       = {Lu, Jianqiao and Zhong, Wanjun and Huang, Wenyong and Wang, Yufei and Zhu, Qi and Mi, Fei and Wang, Baojun and Wang, Weichao and Zeng, Xingshan and Shang, Lifeng and Jiang, Xin and Liu, Qun},
  year         = {2023},
  eprint       = {2310.00533},
  archivePrefix = {arXiv},
  primaryClass = {cs.CL},
  doi          = {10.48550/arXiv.2310.00533}
}

@misc{zhou2024symbolic,
  title        = {Symbolic Learning Enables Self-Evolving Agents},
  author       = {Zhou, Wangchunshu and Ou, Yixin and Ding, Shengwei and Li, Long and Wu, Jialong and Wang, Tiannan and Chen, Jiamin and Wang, Shuai and Xu, Xiaohua and Zhang, Ningyu and Chen, Huajun and Jiang, Yuchen Eleanor},
  year         = {2024},
  eprint       = {2406.18532},
  archivePrefix = {arXiv},
  primaryClass = {cs.CL},
  doi          = {10.48550/arXiv.2406.18532}
}

@misc{wu2026cosplay,
  title        = {Co-Evolving {LLM} Decision and Skill Bank Agents for Long-Horizon Tasks},
  author       = {Wu, Xiyang and Li, Zongxia and Shi, Guangyao and Duffy, Alexander and Marques, Tyler and Olson, Matthew Lyle and Zhou, Tianyi and Manocha, Dinesh},
  year         = {2026},
  eprint       = {2604.20987},
  archivePrefix = {arXiv},
  primaryClass = {cs.AI},
  doi          = {10.48550/arXiv.2604.20987}
}

@misc{xu2026ael,
  title = {{AEL}: Agent Evolving Learning for Open-Ended Environments},
  author       = {Xu, Wujiang and Han, Jiaojiao and Guo, Minghao and Mei, Kai and Zhu, Xi and Zhang, Han and Metaxas, Dimitris N.},
  year         = {2026},
  eprint       = {2604.21725},
  archivePrefix = {arXiv},
  primaryClass = {cs.CL},
  doi          = {10.48550/arXiv.2604.21725},
  url          = {https://arxiv.org/abs/2604.21725}
}

@inproceedings{ALFWorld20,
  title     = {{ALFWorld}: Aligning Text and Embodied Environments for Interactive Learning},
  author    = {Shridhar, Mohit and Yuan, Xingdi and C{\^o}t{\'e}, Marc-Alexandre and Bisk, Yonatan and Trischler, Adam and Hausknecht, Matthew},
  booktitle = {International Conference on Learning Representations},
  year      = {2021},
  url       = {https://openreview.net/forum?id=0IOX0YcCdTn}
}

@article{Gu2025WebDreamer,
  author  = {Yu Gu and Kai Zhang and Yuting Ning and Boyuan Zheng and Boyu Gou and Tianci Xue and Cheng Chang and Sanjari Srivastava and Yanan Xie and Peng Qi and Huan Sun and Yu Su},
  title   = {Is Your {LLM} Secretly a World Model of the Internet? Model-Based Planning for Web Agents},
  journal = {Transactions on Machine Learning Research},
  year    = {2025},
  url     = {https://openreview.net/forum?id=c617yA0HSq}
}

@misc{Cheng2025WebATLAS,
  author        = {Jiali Cheng and Anjishnu Kumar and Roshan Lal and Rishi Rajasekaran and Hani Ramezani and Omar Zia Khan and Oleg Rokhlenko and Sunny Chiu-Webster and Gang Hua and Hadi Amiri},
  title         = {{WebATLAS}: An {LLM} Agent with Experience-Driven Memory and Action Simulation},
  year          = {2025},
  eprint        = {2510.22732},
  archivePrefix = {arXiv},
  primaryClass  = {cs.LG},
  url           = {https://arxiv.org/abs/2510.22732}
}

\clearpage
\FloatBarrier

\appendix
\section{Datasets and Preprocessing}
\label{app:data}
We evaluate our method on four representative agent benchmarks:
ALFWorld\footnote{\url{https://github.com/alfworld/alfworld}}~\citep{ALFWorld20},
ScienceWorld\footnote{\url{https://github.com/allenai/ScienceWorld}}~\citep{Wang2022ScienceWorld},
WebShop\footnote{\url{https://github.com/princeton-nlp/WebShop}}~\citep{yao2022webshop},
and StableToolBench\footnote{\url{https://github.com/THUDM/StableToolBench}}~\citep{guo-etal-2024-stabletoolbench}.
Table~\ref{tab:benchmark_stats} presents task descriptions and data statistics.

\begin{table}[t]
  \centering
  \footnotesize
  \setlength{\tabcolsep}{4pt}
  \renewcommand{\arraystretch}{1.12}
  \begin{tabularx}{\columnwidth}{@{}
    >{\raggedright\arraybackslash}p{1.85cm}
    >{\raggedright\arraybackslash}X
    >{\raggedright\arraybackslash}X
  @{}}
    \toprule
    \textbf{Domain} & \textbf{Task Description} & \textbf{Dataset Statistics} \\
    \midrule

    \multicolumn{3}{c}{\textbf{ALFWorld}} \\
    \midrule
    Household, text-based embodied tasks &
    Six compositional task families: \textsc{Pick}, \textsc{Clean}, \textsc{Heat}, \textsc{Cool}, \textsc{Look}, \textsc{Pick2}. &
    \begin{tabular}[t]{@{}l@{}}
      Training: 3,119 \\
      Test-Seen: 140 \\
      Test-Unseen: 171
    \end{tabular} \\
    \midrule

    \multicolumn{3}{c}{\textbf{ScienceWorld}} \\
    \midrule
    Elementary science curriculum in an interactive text environment &
    30 subtasks with many variations (entities, initial conditions, distractors, room layouts), partitioned following the benchmark protocol. &
    \begin{tabular}[t]{@{}l@{}}
      Training: 1,483 \\
      Test-Seen: 194 \\
      Test-Unseen: 151
    \end{tabular} \\

        \midrule

    \multicolumn{3}{c}{\textbf{WebShop}} \\
    \midrule
    Goal-driven web shopping in a text-based browsing environment &
    Product search, browsing, attribute comparison, and final item selection based on user instructions over noisy webpage observations. &
    \begin{tabular}[t]{@{}l@{}}
      Training: 1,824  \\
      Test: 210
    \end{tabular} \\

    \midrule

    \multicolumn{3}{c}{\textbf{StableToolBench}} \\
    \midrule
    Multi-turn tool-use tasks with executable tool interactions &
    Tool invocation, intermediate execution feedback handling, and solvability-aware evaluation under realistic function-calling constraints. &
    \begin{tabular}[t]{@{}l@{}}
      Training: 1,972 \\
      Test: 169
    \end{tabular} \\
    \bottomrule
  \end{tabularx}
  \caption{Dataset statistics. We report dataset splits following the standard benchmark protocol.}
  \label{tab:benchmark_stats}
\end{table}
\subsection{Datasets}
\paragraph{ALFWorld.}
ALFWorld is a text-based embodied household benchmark in which an agent interacts with a simulated environment through natural-language observations and admissible text actions.
Each episode provides a goal instruction instantiated from compositional task templates, and success requires the agent to perform multi-step planning under partial observability.

\paragraph{ScienceWorld.}
ScienceWorld is a text-based interactive science environment designed to evaluate an agent's ability to solve procedural and reasoning-intensive tasks grounded in everyday scientific phenomena.
Compared with household tasks, ScienceWorld typically involves longer horizons and requires the agent to integrate information gathering, tool use, and multi-step experimentation.

\paragraph{WebShop.}
WebShop is a text-based web shopping benchmark where an agent interacts with simulated e-commerce webpages to find products satisfying a user instruction.
Unlike embodied navigation environments, WebShop exposes the agent to longer and noisier textual observations, including search results, product titles, attributes, and descriptions.
Solving a task requires multi-step browsing, comparison, and selection before making the final purchase decision.

\paragraph{StableToolBench.}
StableToolBench is a multi-turn tool-use benchmark that evaluates an agent's ability to solve tasks by invoking executable tools under realistic interaction constraints.
In contrast to text-navigation environments, StableToolBench emphasizes correct tool selection, robust handling of intermediate execution feedback, and solvability-aware evaluation.
Although its episodes are typically shorter, each decision is more sensitive, since an incorrect tool invocation can immediately derail subsequent progress.
Following existing studies, we report the solvable pass rate (SoPR), which measures average task success on the solvable subset.
Each instance is judged as solved, unsolved, or uncertain, and is mapped to $1$, $0$, or $0.5$, respectively.

\begin{figure*}[t]
    \centering
    \begin{subfigure}[t]{0.48\textwidth}
        \centering
        \includegraphics[width=\linewidth]{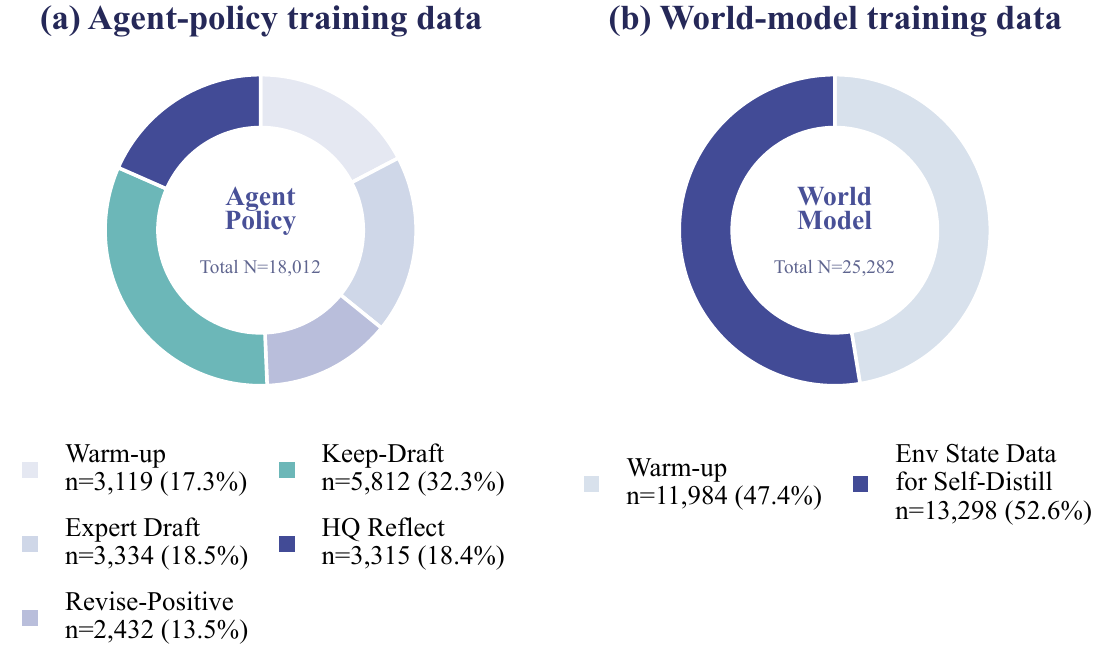}
        \caption{ALFWorld}
        \label{fig:appendix_alfworld_rollout_distribution}
    \end{subfigure}
    \hfill
    \begin{subfigure}[t]{0.48\textwidth}
        \centering
        \includegraphics[width=\linewidth]{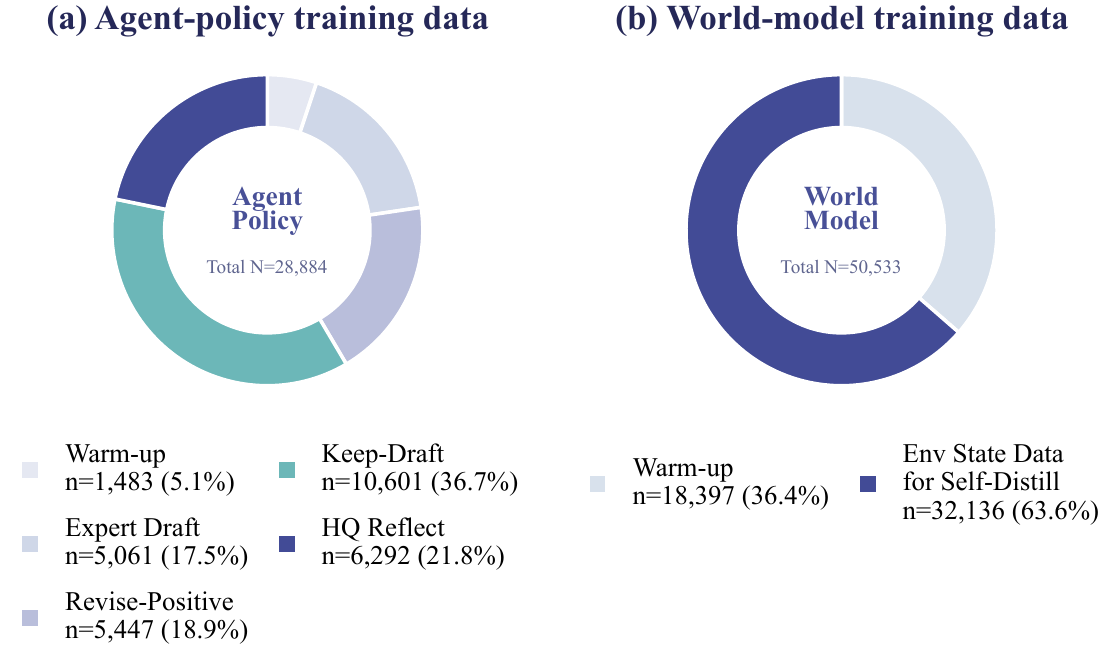}
        \caption{ScienceWorld}
        \label{fig:appendix_scienceworld_rollout_distribution}
    \end{subfigure}

    \vspace{0.8em}

    \begin{subfigure}[t]{0.48\textwidth}
        \centering
        \includegraphics[width=\linewidth]{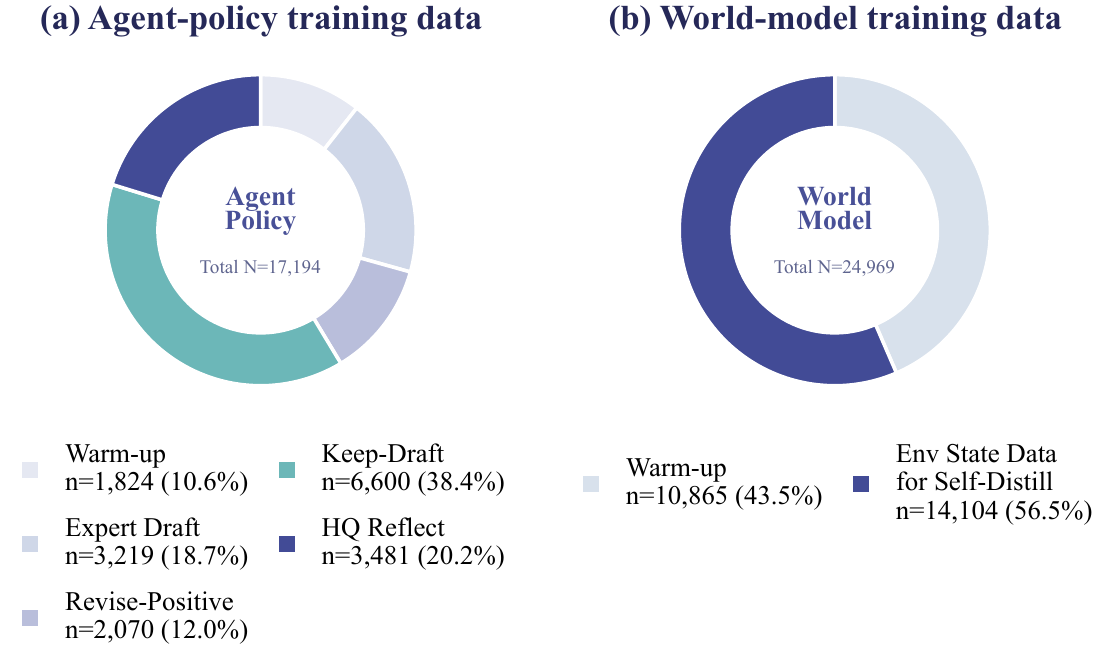}
        \caption{WebShop}
        \label{fig:appendix_webshop_rollout_distribution}
    \end{subfigure}
    \hfill
    \begin{subfigure}[t]{0.48\textwidth}
        \centering
        \includegraphics[width=\linewidth]{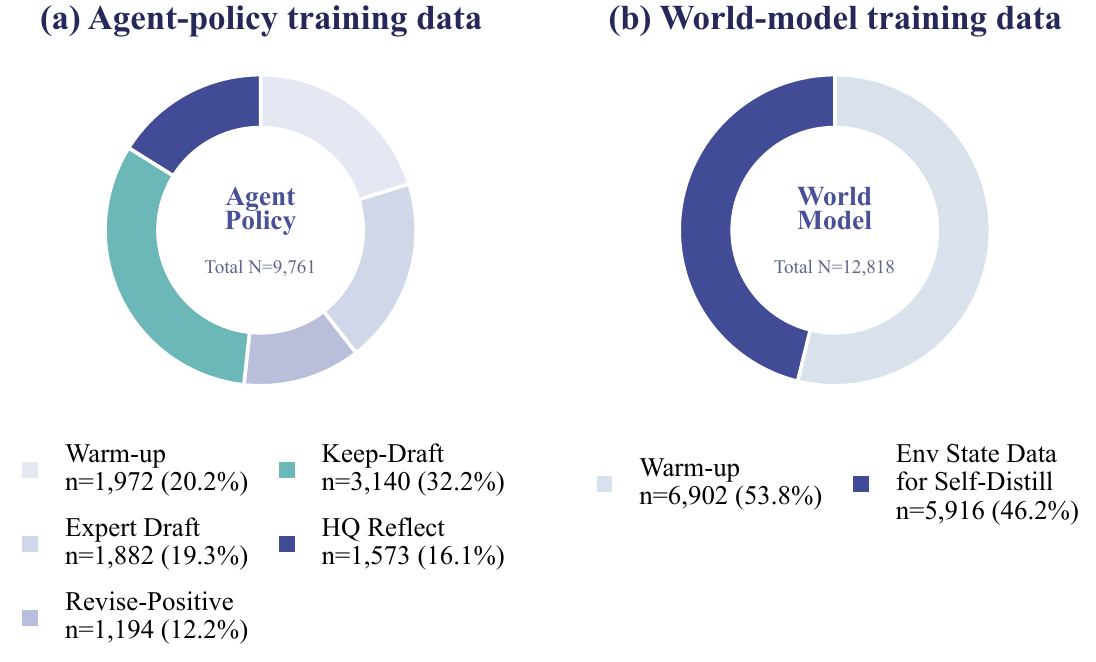}
        \caption{StableToolBench}
        \label{fig:appendix_stabletoolbench_rollout_distribution}
    \end{subfigure}

    \caption{Benchmark-specific rollout training-data distributions. Each panel reports the composition of agent-policy and world-model training data with counts and normalized percentages.}
    \label{fig:appendix_rollout_training_data_distribution}
\end{figure*}
\subsection{Training data}
\label{app:training_data}

Figure~\ref{fig:appendix_rollout_training_data_distribution} summarizes the benchmark-specific training data used by \model.
For the agent policy, the data consists of warm-up demonstrations, expert-draft supervision, revise-positive samples, keep-draft samples, and high-quality reflection samples.
The last three categories are collected from the co-evolving process and provide direct supervision for future-conditioned reflection, i.e., when the policy should revise a draft action and when it should keep the original decision.
For the world model, the data is divided into warm-up transition data and environment-state data for self-distillation.
The warm-up data initializes one-step textual transition modeling, while the self-distillation data comes from on-policy interaction states induced by the evolving agent.
Across benchmarks, a substantial portion of both policy-side and world-model-side data is generated after initialization, showing that \model does not rely only on static offline supervision.
Instead, the policy and world model are continuously refined with interaction-derived signals, which support the closed-loop co-evolving design.

\section{Model Size Scaling}

\begin{figure*}[t]
\centering
\begin{subfigure}[t]{0.38\linewidth}
    \centering
    \includegraphics[width=\linewidth]{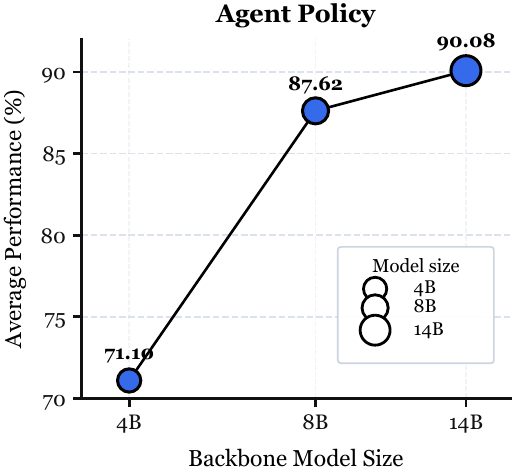}
    \caption{Agent-policy performance scaling.}
    \label{fig:model_size_scaling_agent}
\end{subfigure}
\hfill
\begin{subfigure}[t]{0.38\linewidth}
    \centering
    \includegraphics[width=\linewidth]{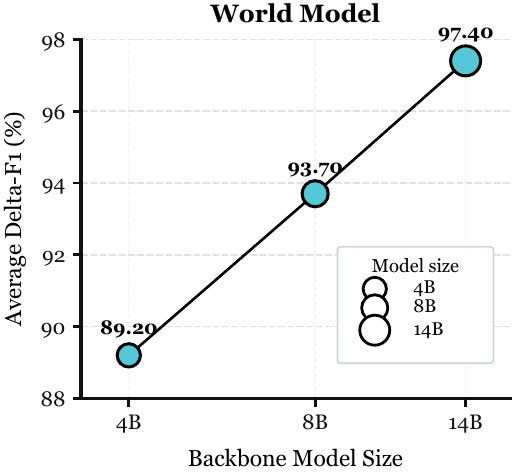}
    \caption{World-model prediction scaling.}
    \label{fig:model_size_scaling_wm}
\end{subfigure}

\caption{Dense-model size scaling of \model. We evaluate \model with dense Qwen3-4B, Qwen3-8B, and Qwen3-14B backbones on ALFWorld. The left panel reports the average agent-policy performance, while the right panel reports the average world-model $\Delta$-F1.
Marker size indicates the relative backbone scale.
}
\label{fig:model_size_scaling}
\end{figure*}

We further study how \model scales with backbone model size by comparing the dense
Qwen3-4B~\citep{qwen3}\footnote{\url{https://huggingface.co/Qwen/Qwen3-4B/blob/main/LICENSE}},
Qwen3-8B~\citep{qwen3}\footnote{\url{https://huggingface.co/Qwen/Qwen3-8B/blob/main/LICENSE}},
and Qwen3-14B~\citep{qwen3}\footnote{\url{https://huggingface.co/Qwen/Qwen3-14B/blob/main/LICENSE}}.
As shown in Figure~\ref{fig:model_size_scaling_agent}, the average agent-policy performance improves monotonically from $71.10$ to $87.62$ and $90.08$ as the backbone scales from 4B to 8B and 14B, indicating that larger policies can better exploit future-conditioned reflection.
Meanwhile, Figure~\ref{fig:model_size_scaling_wm} shows that world-model prediction quality also improves monotonically, with the average $\Delta$-F1 increasing from 89.20\% to 93.70\% and 97.40\%.
These results demonstrate consistent dense-model scaling for both components of \model: the agent policy gains most substantially from 4B to 8B, while the world model continues to exhibit clear improvements up to 14B.

\section{Implementation Details}

All experiments were conducted on a computational cluster equipped with 4$\times$ NVIDIA A100 80GB GPUs.
We report additional implementation details for training cost, parameter settings, and prompting templates as follows.

\begin{table}[t!]
\centering
\scriptsize
\setlength{\tabcolsep}{4.0pt}
\renewcommand{\arraystretch}{1.08}
\resizebox{0.9\linewidth}{!}{
\begin{tabular}{ll}
\toprule
\textbf{Parameter} & \textbf{Value} \\
\midrule

\multicolumn{2}{c}{\textbf{Shared Settings}} \\
\midrule
Maximum sequence length 
& 2048 \\

Optimizer 
& AdamW \\

Learning-rate scheduler 
& Cosine decay \\

Warm-up ratio 
& 0.03 \\

LoRA rank $r$ 
& 8 \\

LoRA scaling $\alpha_{\mathrm{LoRA}}$ 
& 16 \\

LoRA dropout 
& 0.05 \\

\midrule
\multicolumn{2}{c}{\textbf{World Model Training}} \\
\midrule
Training epochs 
& 3 \\

Learning rate 
& $2\times10^{-5}$ \\

Per-device batch size 
& 1 \\

Gradient accumulation steps 
& 16 \\

Co-evolving rounds 
& 3 \\

WMSD weight $\eta$ 
& 0.5 \\

EMA momentum $\mu$ 
& 0.99 \\

World-state threshold $\tau_{\mathrm{wm}}$ 
& 0.6 \\

World-model maximum new tokens 
& 192 \\

\midrule
\multicolumn{2}{c}{\textbf{Agent Policy Training}} \\
\midrule
Training epochs 
& 3 \\

Learning rate 
& $2\times10^{-5}$ \\

Per-device batch size 
& 1 \\

Gradient accumulation steps 
& 16 \\

Co-evolving rounds 
& 3 \\

Online reflection weight $\alpha$ 
& 1.0 \\

BCE weight $\beta$ 
& 0.01 \\

Revise-probability threshold $\tau_p$ 
& 0.5 \\

Reflection-confidence threshold $\tau_q$ 
& 0.6 \\

Draft-action maximum new tokens 
& 64 \\

Reflection maximum new tokens 
& 128 \\

Final-action maximum new tokens 
& 16 \\

Temperature 
& 0.7 \\

Top-$p$ 
& 0.9 \\

\bottomrule
\end{tabular}
}
\caption{
Parameter settings of \model.
We report the default hyperparameters used for the shared backbone, world-model training, agent-policy co-training, and inference decoding.
Here, $\eta$ denotes the WMSD weight in Eq.~(7), while $\alpha$ and $\beta$ denote the weights of online reflection learning and revise-probability learning in Eq.~(8), respectively.
}
\label{tab:parameter_settings}
\end{table}

\subsection{Parameter Settings}
Table~\ref{tab:parameter_settings} reports detailed parameter settings of our proposed \model.

\subsection{Additional Clarification of \model}
\label{app:mechanism_clarification}

This section provides additional details on the decision flow, Future-aware Reflection, world-state gating, teacher-mode supervision, world-model self-distillation, and action gating used in \model.

\paragraph{Draft, refinement, and execution.}
\model separates action generation into three stages: draft generation, future-aware refinement, and final execution.
At step $t$, the action head first proposes a draft action
$a_t^{\mathrm{draft}}$ from the current state $s_t$.
The student world model then predicts an imagined future state
$\hat{s}_{t+1}^{S}=\mathcal{W}_{\phi}^{S}(s_t,a_t^{\mathrm{draft}})$.
Given the gated future context $\bar{s}_{t+1}$, the Future-aware Reflection module produces a refined action
$a_t^{\mathrm{ref}}$ and a refinement probability $p_t$.
The final executed action is determined by the action gate:
$$
a_t^{\mathrm{exec}}
=
\begin{cases}
a_t^{\mathrm{ref}}, & g_t^{\mathrm{act}}=1,\\
a_t^{\mathrm{draft}}, & g_t^{\mathrm{act}}=0.
\end{cases}
$$
Therefore, the environment transition used for subsequent world-model learning is written as
$(s_t,a_t^{\mathrm{exec}},s_{t+1}^{\mathrm{env}})$.
For notational simplicity, the main text may use $a_t$ when the executed action is unambiguous.

\paragraph{Role of textual reflection.}
The Future-aware Reflection module may generate a textual reflection $r_t$ as an intermediate rationale for action refinement.
This rationale helps the policy interpret whether the predicted future state indicates progress, failure, redundancy, or invalidity.
However, $r_t$ is not treated as an independent optimization target.
The supervised outputs of the refinement module are the refined action $a_t^{\mathrm{ref}}$ and the refinement probability $p_t$.
Thus, the reflection text serves as an auxiliary reasoning trace, while the training objective focuses on action-level refinement.

\paragraph{Suffix-based refinement initialization.}
Before online co-evolving, we initialize the Future-aware Reflection module with suffix-based supervision.
The goal is to teach the policy when refinement is necessary, rather than forcing every draft action to be changed.
For each state $s_t$, we compare two suffix branches under the same continuation policy: one branch starts from the draft action $a_t^{\mathrm{draft}}$, and the other starts from the corrected expert action $a_t^{\mathrm{corr}}$.
If the corrected branch achieves a better task outcome or a sufficiently larger suffix return, we set the refinement label $y_t^{\mathrm{ref}}=1$ and use $a_t^{\mathrm{corr}}$ as the target action.
Otherwise, we set $y_t^{\mathrm{ref}}=0$ and keep the draft action as the target.
Equivalent actions after canonicalization are treated as keep-draft cases.
This design prevents the refinement module from over-correcting already useful draft actions.

\paragraph{World-state gate as future-context curriculum.}
The world-state gate controls whether the policy is trained with an imagined world-model state or the real environment state as its future context.
Specifically, the student world model predicts $\hat{s}_{t+1}^{S}$, while the EMA teacher predicts $\hat{s}_{t+1}^{T}$.
We compute the reliability score $\rho_t^{\mathrm{wm}}$ by comparing their action-induced fact changes after canonicalization.
If the student and teacher predictions are consistent, the gate selects the student-predicted state:
$$
\bar{s}_{t+1}=\hat{s}_{t+1}^{S}
\quad \text{if} \quad
g_t^{\mathrm{wm}}=1.
$$
Otherwise, the gate falls back to the real environment state:
$$
\bar{s}_{t+1}=s_{t+1}^{\mathrm{env}}
\quad \text{if} \quad
g_t^{\mathrm{wm}}=0.
$$
This mechanism forms a curriculum over future contexts.
At early stages, when the world model is less reliable, the policy receives cleaner environment states.
As the world model improves, the policy is gradually exposed to imagined future states, better aligning with the inference-time setting.

\paragraph{Teacher mode and privileged information.}
The teacher mode uses the real next environment state $s_{t+1}^{\mathrm{env}}$ only as privileged training-time information for improving the world model.
It is important to distinguish this from inference-time planning.
The privileged state is never provided to the student world model at inference time, nor is it directly exposed to the deployed agent policy.
Instead, the EMA teacher acts as a training-time regularizer that provides soft token-level targets for the student world model.
Therefore, the teacher mode should be understood as a transition-modeling mechanism, not as an oracle planner.

\paragraph{Why world-model self-distillation is used.}
Real-state cross-entropy trains the world model with a single observed next-state sequence.
Although this anchors the model to the actual transition, it provides only one hard target at each decoding position.
World-model self-distillation further transfers the EMA teacher's token-level distribution to the student.
The resulting loss combines real-state anchoring and soft distributional guidance:
$$
\begin{aligned}
\mathcal{L}_{\mathrm{WM}}
=
\mathbb{E}_{\mathcal{D}_{\mathrm{roll}}}
\Big[
&-\log P_{\phi}^{S}
\big(
s_{t+1}^{\mathrm{env}}
\mid
s_t,a_t^{\mathrm{exec}}
\big)
\\
&\quad
+\eta \mathcal{D}_{\mathrm{WMSD}}
\Big].
\end{aligned}
$$
This provides denser supervision over plausible transition descriptions, especially for tokens that correspond to action-induced state changes.
As a result, the student world model can better adapt to the evolving on-policy state-action distribution.

\paragraph{Action gate and refinement confidence.}
The action gate decides whether the refined action should replace the draft action.
It uses three complementary conditions:
$$
\begin{aligned}
g_t^{\mathrm{act}}
=
&\ \mathbf{1}[p_t>\tau_p]\,
\mathbf{1}[q_t^{\mathrm{ref}}>\tau_q]
\\
&\cdot
\mathbf{1}\!\left[
\operatorname{Canon}(a_t^{\mathrm{ref}})
\neq
\operatorname{Canon}(a_t^{\mathrm{draft}})
\right].
\end{aligned}
$$
Here, $p_t$ estimates whether refinement is necessary, while $q_t^{\mathrm{ref}}$ measures whether the generated refined action is reliable.
The canonicalization constraint avoids adopting superficial changes that are semantically equivalent to the draft action.
Together, these conditions reduce unnecessary or low-confidence refinements and make the final execution more conservative.

\paragraph{Evaluation metrics for reflection.}
For each state $s_t$, let $a_t^{\mathrm{draft}}$ and $a_t^{\mathrm{ref}}$ denote the draft and reflected actions, and let $z_t^{\mathrm{draft}},z_t^{\mathrm{ref}}\in\{0,1\}$ be their binary suffix outcomes. We define $m_t=\mathbf{1}[\mathrm{Canon}(a_t^{\mathrm{ref}})\neq \mathrm{Canon}(a_t^{\mathrm{draft}})]$. The metrics are:
$$
\begin{aligned}
\mathrm{URR}
&=
\frac{\sum_t m_t \mathbf{1}[z_t^{\mathrm{draft}}=1]}
{\sum_t \mathbf{1}[z_t^{\mathrm{draft}}=1]+\epsilon},\\
\mathrm{HRR}
&=
\frac{\sum_t m_t \mathbf{1}[z_t^{\mathrm{draft}}=1,z_t^{\mathrm{ref}}=0]}
{\sum_t \mathbf{1}[z_t^{\mathrm{draft}}=1]+\epsilon},\\
\mathrm{BRP}
&=
\frac{\sum_t m_t \mathbf{1}[z_t^{\mathrm{draft}}=0,z_t^{\mathrm{ref}}=1]}
{\sum_t m_t+\epsilon}.
\end{aligned}
$$
URR measures unnecessary revisions, HRR measures harmful revisions, and BRP measures beneficial-revision precision.

\subsection{Training-stage Computational Cost}
We report the training-stage computational cost of \model to improve reproducibility.
Following the main experimental setting, we profile the representative Qwen3-8B configuration on 4$\times$ NVIDIA A100 80GB GPUs.
The training pipeline consists of two phases: warm-up and co-evolving.
The warm-up phase initializes the textual world model with real transition supervision and initializes the agent policy with draft-action imitation and reflect-mode supervision.
The co-evolving phase further updates the world model with real-state supervision and WMSD, while evolving the agent policy through future-conditioned reflection.
GPU-hours are computed as wall-clock time multiplied by the number of GPUs.

\begin{table*}[t]
\centering
\scriptsize
\setlength{\tabcolsep}{5pt}
\renewcommand{\arraystretch}{1.08}
\resizebox{0.95\textwidth}{!}{
\begin{tabular}{lllcc}
\toprule
\textbf{Phase} & \textbf{Module} & \textbf{Main Operation} 
& \textbf{Wall-clock} & \textbf{GPU-hours} \\
\midrule

\multirow{2}{*}{Warm-up}
& World Model 
& Real-transition SFT 
& 2.11 h 
& 8.42 \\

& Agent Policy 
& Draft SFT and reflect-mode initialization 
& 2.84 h 
& 11.36 \\

\cmidrule(lr){2-5}
& \multicolumn{2}{l}{\textit{Subtotal}}
& 4.95 h
& 19.78 \\

\midrule

\multirow{2}{*}{Co-evolving}
& World Model 
& Real-state supervision and world-model self-distillation 
& 2.43 h 
& 9.72 \\

& Agent Policy 
& Future-conditioned reflection evolving 
& 2.05 h 
& 8.18 \\

\cmidrule(lr){2-5}
& \multicolumn{2}{l}{\textit{Subtotal}}
& 4.48 h
& 17.90 \\

\midrule
\textbf{Total}
& \multicolumn{2}{l}{--}
& \textbf{9.42 h}
& \textbf{37.68} \\
\bottomrule
\end{tabular}
}
\caption{
Training-stage computational cost of \model on ALFWorld.
We report wall-clock time and GPU-hours using Qwen3-8B as the backbone model.
The training pipeline is organized into two phases: warm-up and co-evolving.
Each phase is further decomposed into the world-model side and the agent-policy side.
}
\label{tab:comap_training_cost}
\end{table*}

In the initialization phase, the world model and the agent policy are trained with lightweight supervised objectives.
The world model learns the one-step textual transition distribution from real environment transitions, while the draft mode of the policy is anchored by expert action imitation.
We also initialize the reflection mode with trajectory-suffix supervision so that the policy can learn when a draft action should be revised and what corrected action should be produced.

In the co-evolving phase, the additional cost comes from the closed-loop optimization of the two modules.
On the world-model side, \model performs on-policy self-distillation, where the student world model is trained with real next-state supervision and token-level soft guidance from an EMA teacher.
On the policy side, \model collects future-conditioned reflection samples and updates only high-quality online corrections selected by the world-state gate and the action gate.
Although this phase introduces extra computation compared with ordinary supervised fine-tuning, the overall cost remains moderate because \model uses one-step imagination rather than multi-branch tree search or long-horizon rollout expansion.

At inference time, \model removes the EMA teacher and uses only the student world model together with the agent policy.
Therefore, the self-distillation teacher introduces training-time overhead but does not increase the number of deployed models at test time.

\subsection{Details on API-called Model Evaluation}
\label{app:api_model_details}

For reproducibility, GPT-5.4 is evaluated with the official API model identifier \texttt{gpt-5.4}~\citep{openai2026gpt54}, and the DeepSeek baselines are evaluated with \texttt{deepseek-v4-pro} and \texttt{deepseek-v4-flash}~\citep{deepseek2026v4}. 
All API-called models use the same ReAct prompt, decoding configuration, and text-only benchmark interface, without browser, computer use, file search, retrieval, or external tool access beyond the benchmark-provided actions.
We use OpenAI GPT-5.4 with the documented API alias 
\texttt{gpt5.4}\footnote{\url{https://platform.openai.com/docs/models/gpt-5.4}}.
For DeepSeek baselines, we use \texttt{deepseek-v4-pro} and 
\texttt{deepseek-v4-flash}\footnote{\url{https://api-docs.deepseek.com/api/list-models}}.

\subsection{Multi-step Prediction and Compounding Error}
\label{app:multistep}

We evaluate recursive lookahead on ScienceWorld with Qwen3-4B. For policy evaluation, only the first predicted action is executed before replanning from the real observation. Performance decreases with the prediction horizon for both methods.
\model mitigates but does not eliminate compounding error; therefore, we use one-step prediction for the best accuracy--efficiency trade-off.

\begin{table}[t]
\centering
\small
\begin{tabular}{llcccc}
\toprule
Checkpoint & Metric (\%) & $H{=}1$ & $H{=}2$ & $H{=}5$ & $H{=}10$ \\
\midrule
SFT & $\Delta$F1 & 44.37 & 34.92 & 18.46 & 7.84 \\
SFT &  SR & 39.35 & 36.81 & 26.94 & 12.66 \\
\model & $\Delta$F1 & 86.08 & 76.45 & 55.36 & 32.41 \\
\model & SR & 51.57 & 49.28 & 39.68 & 24.81 \\
\bottomrule
\end{tabular}
\caption{Multi-step prediction ($\Delta$F1) and policy performance (success rate, SR) on ScienceWorld.}
\label{tab:multistep}
\end{table}

\subsection{State-restoration Requirement}
\label{app:state_restoration}

Reflection-mode initialization assumes that the environment can be restored to an earlier state so that alternative action branches can be compared.
Consequently, this procedure is most directly applicable to simulated or
sandboxed environments. This requirement applies only during offline
initialization: \model neither restores nor branches the environment at
inference time and executes only the gated action. 
Applying the same initialization procedure to live websites or stateful APIs with irreversible side effects would require reset-free supervision derived from forward-only interaction logs, which we leave for future work.

\subsection{Prompting Templates}
We provide three prompt templates for \model in Figure~\ref{fig:prompt_draft_action}, Figure~\ref{fig:prompt_reflection_future_state}, and Figure~\ref{fig:prompt_world_model}. 
The agent policy uses two prompts: one for generating a draft action and one for reflecting on the draft action with a future-state signal. 
The world model uses a separate prompt to predict the next state after a given action. 
This design separates policy decision-making from one-step transition prediction.
\begin{figure*}[t]
    \centering
    \begin{tcolorbox}[
        title=\textbf{Prompt Template for Draft Action Generation},
        colframe=blue!50!black,
        colback=blue!5!white,
        coltitle=white,
        fonttitle=\bfseries,
        arc=1mm,
        boxsep=2pt,
        fontupper=\small
    ]
        You are an intelligent agent interacting with a text-based environment.
        Your goal is to complete the task by taking valid and useful actions.

        At the current step, you will be given the task goal, the current observation, the previous action, and the latest environment feedback.
        You should understand the current situation and propose one candidate action for the next step.

        \vspace{4pt}
        \textbf{Instructions:}
        \begin{enumerate}[nosep, leftmargin=*]
            \item Briefly summarize the current state and the latest feedback.
            \item Identify what has already been achieved and what remains to be done.
            \item Choose the next action that is most likely to make progress toward the task goal.
            \item Output exactly one candidate action.
        \end{enumerate}

        \vspace{4pt}
        \textbf{Response format:}
        \begin{tcolorbox}[colback=white, colframe=gray!30, arc=0mm, boxsep=0pt,
            left=4pt, right=4pt, top=2pt, bottom=2pt]
            \ttfamily\footnotesize
            Reason: briefly describe the current situation and relevant feedback\\
            Thought: briefly explain the next-step plan\\
            Draft Action: output exactly one executable action
        \end{tcolorbox}

        \vspace{4pt}
        \textbf{Inputs:}
        \begin{tcolorbox}[colback=white, colframe=gray!30, arc=0mm, boxsep=0pt,
            left=4pt, right=4pt, top=2pt, bottom=2pt]
            \ttfamily\footnotesize
            Task goal:\par
            \{task\_goal\}\par\medskip

            Current observation:\par
            \{current\_observation\}\par\medskip

            Previous action:\par
            \{previous\_action\}\par\medskip

            Latest environment feedback:\par
            \{latest\_feedback\}\par\medskip

            Interaction history:\par
            \{history\}
        \end{tcolorbox}

        \vspace{4pt}
        \textbf{Rules:}
        \begin{itemize}[nosep, leftmargin=*]
            \item The action must be valid in the current environment.
            \item Do not invent unavailable objects, locations, webpages, tools, or arguments.
            \item If the latest feedback indicates that the previous action failed, avoid repeating the same failed action.
            \item Do not output multiple candidate actions.
        \end{itemize}
    \end{tcolorbox}
    \caption{
    Prompt template for draft action generation in the agent policy.
    The policy proposes a candidate action from the current task context before querying the world model.
    }
    \label{fig:prompt_draft_action}
\end{figure*}

\begin{figure*}[t]
    \centering
    \begin{tcolorbox}[
        title=\textbf{Prompt Template for Reflection with Future State},
        colframe=blue!50!black,
        colback=blue!5!white,
        coltitle=white,
        fonttitle=\bfseries,
        arc=1mm,
        boxsep=2pt,
        fontupper=\small
    ]
        You are an intelligent agent revising a candidate action using a future-state signal.
        The future-state signal describes the possible consequence of executing the draft action.
        Your goal is to decide whether the draft action should be kept or revised.

        Revise the draft action only when the future-state signal shows that the draft action is invalid, unhelpful, harmful, redundant, or clearly worse than another available action.
        If the draft action already makes useful progress, keep it unchanged.

        \vspace{4pt}
        \textbf{Instructions:}
        \begin{enumerate}[nosep, leftmargin=*]
            \item Read the task goal, current observation, draft action, and future-state signal.
            \item Judge whether the future-state signal indicates progress, failure, no change, or a harmful consequence.
            \item Decide whether to keep or revise the draft action.
            \item Output exactly one final action for execution.
            \item Output a revise probability to indicate how strongly the draft action should be revised.
        \end{enumerate}

        \vspace{4pt}
        \textbf{Response format:}
        \begin{tcolorbox}[colback=white, colframe=gray!30, arc=0mm, boxsep=0pt,
            left=4pt, right=4pt, top=2pt, bottom=2pt]
            \ttfamily\footnotesize
            Reflection: analyze whether the draft action leads to useful progress\\
            Decision: KEEP or REVISE\\
            Final Action: output exactly one executable action\\
            Revise Probability: output a number from zero to one
        \end{tcolorbox}

        \vspace{4pt}
        \textbf{Inputs:}
        \begin{tcolorbox}[colback=white, colframe=gray!30, arc=0mm, boxsep=0pt,
            left=4pt, right=4pt, top=2pt, bottom=2pt]
            \ttfamily\footnotesize
            Task goal:\par
            \{task\_goal\}\par\medskip

            Current observation:\par
            \{current\_observation\}\par\medskip

            Draft action:\par
            \{draft\_action\}\par\medskip

            Future-state signal:\par
            \{future\_state\}\par\medskip

            Interaction history:\par
            \{history\}
        \end{tcolorbox}

        \vspace{4pt}
        \textbf{Rules:}
        \begin{itemize}[nosep, leftmargin=*]
            \item If the draft action should be kept, repeat it as the final action.
            \item If the draft action should be revised, output only the revised action as the final action.
            \item Do not revise merely for stylistic reasons.
            \item Do not invent new objects, locations, webpages, tools, or arguments.
            \item The final action must be directly executable in the environment.
        \end{itemize}
    \end{tcolorbox}
    \caption{
    Prompt template for reflection with future state in the agent policy.
    The policy uses the future-state signal to decide whether to keep or revise the draft action.
    }
    \label{fig:prompt_reflection_future_state}
\end{figure*}

\begin{figure*}[t]
    \centering
    \begin{tcolorbox}[
        title=\textbf{Prompt Template for Textual World Model},
        colframe=green!45!black,
        colback=green!5!white,
        coltitle=white,
        fonttitle=\bfseries,
        arc=1mm,
        boxsep=2pt,
        fontupper=\small
    ]
        You are a textual world model for an interactive agent environment.
        Given the current textual state and an action, predict the next textual state after executing the action.

        \vspace{4pt}
        \textbf{Input:}
        \begin{tcolorbox}[colback=white, colframe=gray!30, arc=0mm, boxsep=0pt,
            left=4pt, right=4pt, top=2pt, bottom=2pt]
            \ttfamily\footnotesize
            Current state:\par
            \{state\}\par\medskip

            Action:\par
            \{action\}
        \end{tcolorbox}

        \vspace{4pt}
        \textbf{Instruction:}
        Predict only the one-step next state caused by the given action.
        Preserve unchanged facts and update only the parts directly affected by the action.
        Do not generate a new action, plan, explanation, or multi-step rollout.

        \vspace{4pt}
        \textbf{Output format:}
        \begin{tcolorbox}[colback=white, colframe=gray!30, arc=0mm, boxsep=0pt,
            left=4pt, right=4pt, top=2pt, bottom=2pt]
            \ttfamily\footnotesize
            Predicted next state: \{next\_state\}
        \end{tcolorbox}
    \end{tcolorbox}
    \caption{
    Prompt template for the one-step textual world model.
    Given the current textual state and an action, the world model predicts the resulting next state.
    }
    \label{fig:prompt_world_model}
\end{figure*}

\end{document}